\documentclass[acmsmall]{acmart}
\usepackage[ruled]{algorithm2e}
\usepackage{subcaption}

\AtBeginDocument{%
  \providecommand\BibTeX{{%
    \normalfont B\kern-0.5em{\scshape i\kern-0.25em b}\kern-0.8em\TeX}}}

\setcopyright{acmcopyright}
\copyrightyear{2019}
\acmYear{2019}

\acmJournal{THRI}

\begin{document}


\title{Mutual Reinforcement Learning}

\author{Sayanti Roy}
\affiliation{%
  \department{Department of Computer Science}
  \institution{Oklahoma State University}
  \streetaddress{219 MSCS}
  \city{Stillwater}
  \state{OK}
  \postcode{74078}
  \country{USA}}
\email{sayanti.roy@okstate.edu}
\author{Emily Kieson, Charles Abramson}
\affiliation{%
  \department{Department of Pshycology}
  \institution{Oklahoma State University}
  \streetaddress{116 N Murray Hall}
  \city{Stillwater}
  \state{OK}
  \postcode{74078}
  \country{USA}}
\email{kieson@okstate.edu}
\email{charles.abramson@okstate.edu}
\author{Christopher Crick}
\affiliation{%
  \department{Department of Computer Science}
  \institution{Oklahoma State University}
  \streetaddress{219 MSCS}
  \city{Stillwater}
  \state{OK}
  \postcode{74078}
  \country{USA}}
\email{chriscrick@cs.okstate.edu}


\begin{abstract}
Recently, collaborative robots have begun to train humans to achieve
complex tasks, and the mutual information exchange between them can
lead to successful robot-human collaborations. In this paper we
demonstrate the application and effectiveness of a new approach
called \textit{mutual reinforcement learning} (MRL), where both humans
and autonomous agents act as reinforcement learners in a skill transfer scenario
over continuous communication and feedback. An autonomous agent
initially acts as an instructor who can teach a novice human
participant complex skills using the MRL strategy. While teaching
skills in a physical (block-building) ($n=34$) or simulated (Tetris)
environment ($n=31$), the expert tries to identify appropriate reward channels
preferred by each individual and adapts itself accordingly using an
exploration-exploitation strategy. These reward channel preferences
can identify important behaviors of the human participants, because
they may well exercise the same behaviors in similar situations
later. In this way, skill transfer takes place between an expert
system and a novice human operator. We divided the subject population
into three groups and observed the skill transfer phenomenon,
analyzing it with Simpson"s psychometric model.  5-point Likert scales
were also used to identify the cognitive models of the human
participants. We obtained a shared cognitive model which not only
improves human cognition but enhances the robot's cognitive strategy
to understand the mental model of its human partners while building a
successful robot-human collaborative framework.
\end{abstract}

\ccsdesc[500]{Robot Human Collaboration~Mutual Reinforcement Learning(MRL)}
\ccsdesc[300]{Robot Human Collaboration~Robot-Human skill transfer}
\ccsdesc{Computer Human Interaction~MRL}
\ccsdesc[100]{Computer Human Interaction~skill transfer}


\maketitle

\begin{figure}
  \includegraphics[width=10cm,height=7cm]{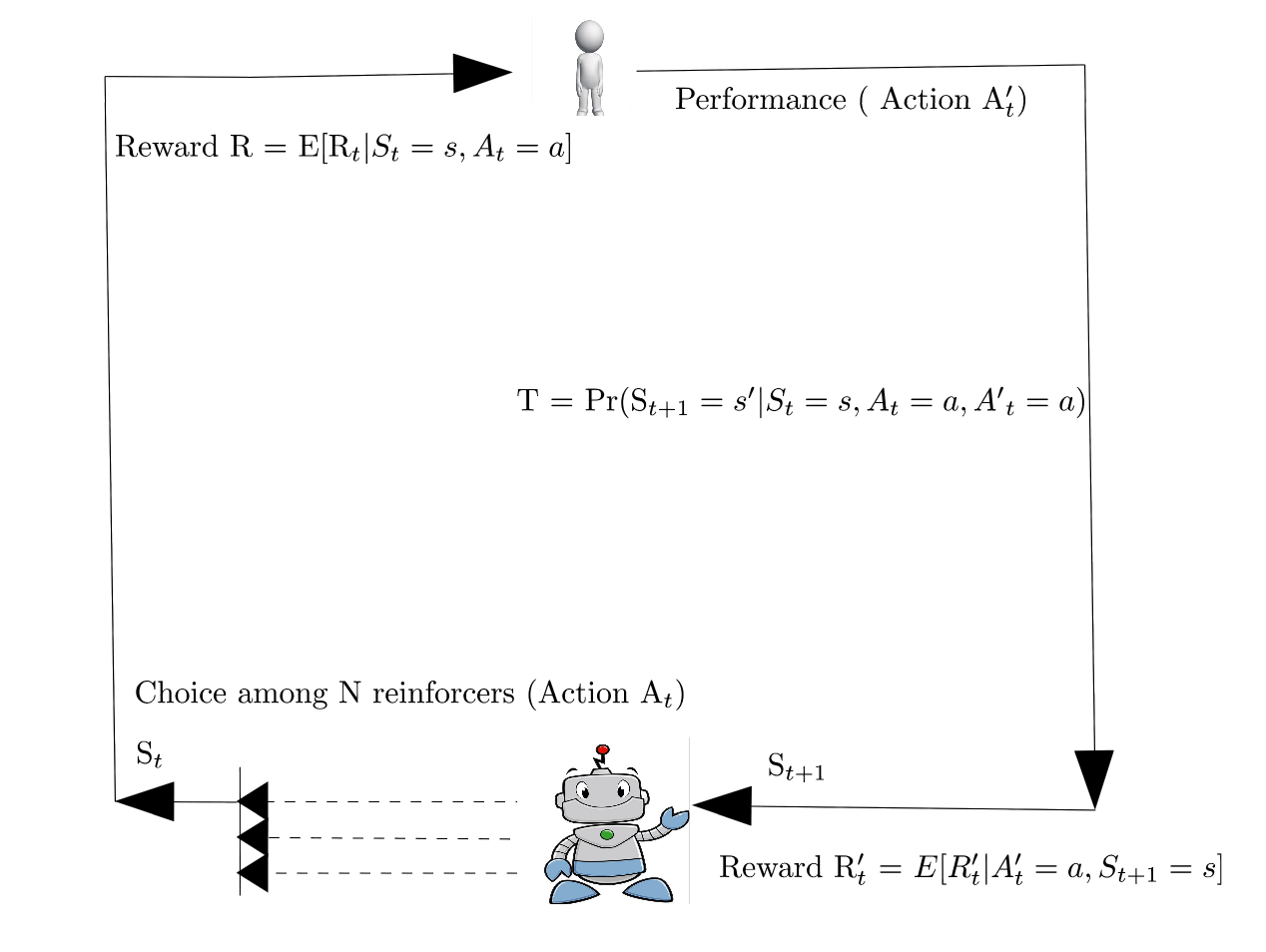}
  \caption{Mutual Reinforcement Learning illustrated.  Robots and
    humans are both acting as reinforcement learners and trying to
    learn from each other to accomplish a task.  The action $A_t$ of
    the robot is acting as a reward $R_t$ to the human in state $S_t$,
    while the action of the human $A^\prime_{t}$ is acting as a reward
    $R^\prime_{t}$ to the robot, changing the state to $S_{t+1}$.}
  \label{fig:one}
\end{figure}

\section{Introduction} 


Daily experiences influence our learning and change the way we think
and act. Sometimes we are not even aware that we are learning from our
surroundings, which is a very informal way of perceiving things. On
the other hand, we can also learn in a formal way from a a structured
classroom environment. Learning is not limited to acquiring knowledge
or facts; we also learn skills and attitudes. This can happen in
different ways. We learn new ideas and concepts from a lecture or a
discussion, whereas skills must be acquired via continuous practice
and receiving simultaneous feedback from an instructor. In a planned
environment, learning is reinforced by teachers who expect students to
memorize the content and later reward them for it.  In contrast,
researchers or scientists learn by investigating things themselves,
over time. But in any form of learning, motivations and rewards play a
very important role, as people derive satisfaction from the feeling of
competence.  In the case of learning a new skill, people can be
strongly motivated by the incentives they are given, which might lead
them to acquire new knowledge which they can use in future.

Teachers are entrusted with the job of determining what the student
will learn.  They are not only instrumental to the students' learning,
but also make sure that they have learnt the subject properly.
Teaching must be planned very carefully, taking the learning styles
and the background of the students into account. Teachers also need to
assess students often to determine how well they are progressing and
simultaneously attend to their weaknesses.  Hence teaching and
learning are constructed over a series of intrinsic and extrinsic
social interactions which influence the cognitive models of both the
teacher and the learner.  The learning can be improved if the
facilitator can teach each individual, possessing some understanding
of the subtleties of the student's mind, behavior and learning style,
and regulating the motivational strategies accordingly
\cite{prozesky2000teaching}.

\subsection{Contributions}

In this research, the humanoid robot Baxter and a computer system
(depending on the experiment) motivate an individual extrinsically
during the learning process using several positive reinforcers. During
the interactions, the robot or the system initially demonstrates the
task to the participant, and then provides learned reinforcers to make
sure the skill is transferring properly, using Simpson's psychometric
model \cite{simpson1972classification}, and concurrently learns about
their cognitive models. The expert uses a reinforcement learning
strategy to understand the effect of its reinforcement presentation on
its human subjects, attempting to increase their performance over time
\cite{roy2019mutual,royeffective,roy2018using}.  To identify the
success of MRL-guided skill transfer, we divided the subject
population into three groups where participants get no reinforcement,
random reinforcement, or individually-tailored learned reinforcement
(MRL) respectively. We compared the number of mistakes in each group,
because we expect MRL to be more effective for subjects who are
performing somewhat poorly.  We discovered that participants in the
learned group were more likely to perform well and committed
comparatively fewer mistakes with respect to the other experimental
conditions ($p<0.05$ for Baxter) and ($p<0.05$ in random group for the computer-based
Tetris skill learning).  We also determined information gain over time
and how a machine's regret strongly correlates with the probability
that a test subject makes more versus fewer mistakes. In addition, we
produced confusion matrices demonstrating the effectiveness of MRL in
the experiments using 5-point Likert scale data.


\section{Related work}

Many contemporary researchers are working on identifying appropriate
reward channels in human-robot collaborative frameworks to maximize
performance. The following section briefly discusses this work along
with the influence of positive reinforcers on motivating humans in
skill transfer procedures.

\subsection{Reinforcement learning techniques to identify better reward channels}

Rewards play a crucial role in both identifying and shaping a person's
behavior. They not only tell us about a person's personality, but
provide an influence channel when used effectively.  Hence, recently
many scientists are interested in researching appropriate reward
channels that might increase task performance. Lopes
\cite{Lopes2013multi} upgraded Multi-Arm Bandit techniques using
different motivational resources to maximize skills and learning
activities. He also researched the recovery of reward functions
$\mathbb{R}_{xa}(p)$ from expert demonstrated policies $\pi_{i}$
\cite{Lopes-ECML-09} to ensure active learning. The modification of
traditional reinforcement learning algorithms using reward shaping
produced important insights into how skill and accuracy can be
improved for a particular task. Cooperative inverse reinforcement
learning (CIRL) uses a human reward function $ R:S \times
{\mathbb{A}^H} \times {\mathbb{A}^R} \times \Theta \longrightarrow
\mathbb{R} $ that maps world states, joint actions, and reward
parameters to real numbers to establish useful human robot
collaboration, where the robot is unaware of the initial reward. CIRL
can be used in various platforms like active teaching, active
learning, and communicative actions that are more effective in
achieving value alignment \cite{hadfield2016cooperative}. These
researchers are also using multi-arm bandit techniques (MAB) to
address the problems with assistive agents who can help human
participants to select appropriate channels to maximize the cumulative
reward \cite{chan2019assistive}. Here the human does not know the
reward function but can learn it through several interactions, whereas
the robot only observes the human interactions and not the reward
associated with it.  Tabrez et al. used their Reward Augmentation and
Repair through Explanation (RARE) framework for estimating task
understanding where the autonomous agent detects potential causes of
system failures and uses human-interpretable feedback for model
correction \cite{tabrez2019explanation}. Nikolaidas et
al.\cite{nikolaidis2013human} described a human-robot cross-training
framework using reinforcement learning techniques where humans and
robots switch roles to improve the overall performance. Li et al. used
MRL in automatic poetry generation using two models (local and global)
which have some predefined criteria as rewards, and they learn from
each other to pursue higher scores \cite{yi2018automatic}. Griffith
\cite{griffith2013policy} discussed novel policy shaping algorithms
and how motivations and reward signals can be used as a channel to
impact human-robot partnerships in an HRI setting, simultaneously
improving the future learning process of both humans and robots. Knox
et al. \cite{Knox:2009:ISA:1597735.1597738, 4640845} designed a novel
framework named TAMER which allows a human to train a learning agent
to perform a complex tasks over continuous interaction. In our
previous papers \cite{roy2018using} we have also discussed how the
robot updates its own cognitive model with each human interaction,
improving the overall task performance through
exploration-exploitation strategies. In this work we are not only
extending the goal of our previous work, but also updating the MRL
technique for better results.

\subsection{Robots learning to teach}

Many scientists have started exploring this new area of robotics,
where along with robot teaching we can gain other useful information
about robot and human behaviours from their interaction. Spaulding
\cite{spaulding2018social} introduced an integrated system for
autonomously analyzing and assessing children's speech and
pronunciation in the context of an interactive word game between a
social robot and a child.  This approach used Gaussian Process
Regression (GPR), augmented with an active Learning protocol that
informed the robot's behavior. Scassellati
\cite{ramachandran2015developing,litoiu2015robotic} and Park
\cite{park2017growing} have presented feedback-based human-robot
interaction, demonstrating that if humans are guided by a robot at an
interpersonal level, it increases the robot's perceived social
reliability and makes humans more eager to interact with it. A robot
learning from human feedback tends develop a mental model
\cite{lee2005human,scassellati2001foundations} of its own which
positively influences human cognition. Fasola et
al. \cite{Fasola:2013:SAR:3109708.3109710} used socially assistive
robots (SAR) to train elderly humans in physical fitness by motivating
them. Yin et al. \cite{yin2015bidirectional} described intelligent
robot systems acquiring human-like writing style and then exploiting
it to teach children. Fan et al. \cite{fan2018learning} used neural
network models to evaluate teaching strategies when one intelligent
system is trying to teach another.  Leite et al.  \cite{leite2012long}
used robots to socially support children in a game scenario.  The
robot not only increased the performance accuracy of the human
learner, but also connected with them emotionally and provided social
assistance throughout their learning process. This social support
helped the children to build their self esteem and encouraged them to
perform better. Cakmak \cite{cakmak2009effects} demonstrated how
social learning strategies vary with the particular environment when
robots are allowed to explore and learn from their surroundings.

In this paper, along with the effectiveness of MRL, we are also
concerned with the idea of robots learning to be good teachers.  We
use a robot's own predicted regret and confusion matrices to evaluate
its own cognitive model.

\subsection{Empathy and positive reinforcers}

Empathy is based in the social-cognitive and behavioral ability to
vicariously experience another person or animal's affect, and is
critical in the social interactions of humans and some animals
\cite{keskin2014isn,lockwood2016anatomy}. Empathy plays a vital role
in social interaction in all stages of human life and many
contemporary researchers are working on empathetic robots that are
designed to respond to human behavior and emotion with appropriate
social cues.  Empathy and adaptation may not be enough, however, since
social responses are only one component of effective human-robot
interactions. Instead, robot interactions that facilitate mutual
learning with the human counterpart may prove more effective in a
teaching environment due to the ability to learn, adapt, and create
reinforcement feedback tailored to the individual. Hence the ability
to empathize has also been found to be a critical characteristic of
effective teachers. In one study, teachers demonstrating more empathy
were able to adapt the structure, behavior, and manifestation of
empathy based on the group or individual and provide more effective
teaching strategies \cite{Mihaela2013}.  In our research, we used
several positive reinforcers as reward channels to interact with the
human participant.

In order to effectively study human-robot interactions and learning,
scientists have incorporated other socially-inspired tools in addition
to empathy.  For example, auditory and visual cues are important in
learning exchanges between humans and robots, especially when learning
through demonstration \cite{Koenig2010}. Furthermore, modeling
demonstration learning using robots and humans has shown to be
effective and the closer the demonstration technique was to typical
social learning, the more rapport the participant felt with the robot
and the more he or she learned \cite{Sauppe:2015:ETT:2825762.2825778}.
Humans also demonstrate a need to share intentions with their social
partners, and in order to mimic this with robots, the robot partner
needs to mimic the social skills necessary to interact with humans and
demonstrate shared intention \cite{Dominey2011}.  The robot, in this
case, demonstrated the ability to learn a goal and intentional actions
linked to the goal through cooperative learning \cite{Dominey2011}. In
these cases, behavioral interactions and social acceptance are
critical components to the human-robot interaction.

It is possible for humans to respond to perceived empathy from robot
and computer interactions.  Research shows that individuals perceive
empathy through digital devices and computer-mediated interactions
\cite{Powell2016}, and additional studies are developing robot-human
interactions that more closely mimic human-human interactions using
touch and visual interactions \cite{Salter2006}.  Furthermore, empathy
increases rapport between humans and robots, which is important for
user comfort \cite{article2012}. This suggests that, while empathy is
important for contextual comfort, it may not be the only component of
a learning environment and does not indicate a human response for the
robot.  While scientists may have developed robots to mimic empathy
that can be detected by participants, humans have yet to respond with
equal attachment or empathy towards robots \cite{Konok2018}.  In other
words, while adaptive empathetic robots may build some rapport with
humans, the communication is only from the robot to the human; the
robot is not necessarily responding in ways that may be necessary for
human learning.  A few researchers have explored various areas where
positive reinforcement from robots had a large impact on children.
Boccanfuso et al. \cite{boccanfuso2016emotional} investigated the
difference in responses between children with or without autism with
an emotion-stimulating robot using positive reinforcement in an
interactive environment. Nunez et al. \cite{nunez2015humanoid}
described the use of positive reinforcers to overcome the underlying
challenges in motivating a child to continue learning and to share the
experience with others. Kim \cite{kim2013social} addressed the unique
positive effects and advantages a robot can have on autistic children,
exploring areas where robots play an important role in the lives of
specific individuals.  We wish to investigate how a robot can develop
an understanding of the underlying motivations and cognitive traits of
individual people, so that it can shape its teaching strategies
appropriately and enhance the learning process.
 
\subsection {Positive reinforcers in MRL}

Mutual feedback between robot and human has become increasingly
important in human-robot interactions.  Interactivism
\cite{bickhard2009interactivism} and process-oriented robots have been
challenging in the past since there is a necessary balance between
environmental stimuli and feedback and the adaptation of software and
processes that can adapt and change with them \cite{Stojanov2006}.
Robots using socially-inspired reinforcement including verbal and
behavioral feedback have only shown modest results, and studies have
suggested that a more targeted approach tailored to the individual
would be better suited for future robot-human interactions
\cite{Ferreira:2015:RBD:2794083.2794125}.

To better facilitate natural social interactions and engage with the
learning environments of humans, robots need to adapt and respond
appropriately to each individual.  Positive reinforcement increases
learning in all animals and promotes voluntary behaviors of animals,
but the reinforcement tools need to be species-specific and based on
individual preferences and experiences.  In this sense, if robot-human
interactions are to use socially-derived reinforcements as teaching
tools, researchers need to take into account not just human social
interactions, but individual differences as well. This means that the
robots need to be programed with an understanding of
individual-specific approaches to interactions based on principles of
learning and be able to adapt and respond in ways that are tailored
based on the individual's unique responses.

The scientists in this study have developed a novel approach using
mutual reinforcement learning where both the robot and human act as
individual empathizers who can act as reinforcement learning agents to
achieve a particular task. Thus in this paper the humanoid robot
Baxter and a personal computer not only adapts or empathizes with its
human participant but also takes a step forward to encourage them and
achieve their goal.



\section {Technical description} 


Mutual reinforcement learning (MRL) deals with the scenario where both
humans and autonomous agents act as reinforcement learners for each
other, identifying the path to achieve maximum reward.  In this
instance, the robot initially acts as an expert and its human
counterpart as a novice.  In MRL, one agent's action is mapped as a
reward to another.  Here the agents, as they are unaware of each
other's incoming actions, discover the appropriate reward channel over
continuous communications with one another.  The autonomous agent
acting as an expert learns about the appropriate reward channel
through an exploration-exploitation tradeoff
\cite{audibert2009exploration}.  The action of the novice agent
(judged in terms of the agent's performance) not only affects the
immediate rewards but also the expert's next action.  The expert does
not immediately jump to a conclusion about the decision to be made,
but rather invests more time and effort in accumulating further
information, with the hope that a broader perspective will lead to a
better decision in the future.  On the other hand, the humans
interpret the actions of the robot (or computer) agent as a reward,
which influences their performance in learning the task. In this
paper, MRL is implemented in a skill transfer scenario, where the
autonomous agent is trying to teach a human some complex task, while
updating its own mental model \cite{sutton2018reinforcement} at the
same time.

MRL is a tuple $\{S,A(A^\prime),T,R(R^\prime)\}$ where $S$ is a set of
states; $A$ and $A^\prime$ are sets of actions; $T$ is the set of
state transition probabilities $p(s,a)$ upon taking action $a$
$\epsilon$ $A$ or $A^\prime$ in state $s$, and $r$ $\epsilon$ $R$ and
$R^\prime : (S,A)$ or $(S,A^\prime) \Rightarrow R$ and $R^\prime$ are
the reward functions. Since, in MRL, the action of an agent is the
reward to another and vice versa, the tuple can be simplified as
follows: Novice=$\{S,A^\prime,T,R\}$, Expert=$\{S,A,T,R^\prime\}$
where if the novice executes action $A^\prime$, reward $R^\prime$ is
received by the expert.  This helps the expert to execute action $A$
using an exploration/exploitation strategy, which at the same time
acts as a reward $R$ to the participant. If the action $A^\prime$ is
successful, then the robot realizes that the participant is fonder of
reward $R$, which acts at the same time as reward $R^\prime$ for the
robot to understand its own performance or action $A$. Here the reward
for the novice $r$ $\epsilon$ ${(r_1,r_2,r_3...)}$ is selected by an
exploration-exploitation tradeoff where $r_1 , r_2, r_3$ are all
different kinds of reinforcers mentioned in Section
\ref{reinforcers}. In the case of MRL, we have a verbal, hint, gesture
and simple feedback for the robot whereas there are seven different
reinforcers in case of Tetris. Therefore the expected rewards in both
the cases in a state action pair can be written as a two-argument
function $ r : S X A \Rightarrow r \epsilon R $ and $ r : A^\prime X S
\Rightarrow r \epsilon R^\prime $.

\begin{equation}
r(s,a) \doteq E[R_t | S_{t} = s, A_{t} = a ] \label{eq:1}
\end{equation}

\begin{equation}
r(s,a) \doteq E[R^\prime_{{t}} | A^\prime_{t} = a, S_{t+1} = s] \label{eq:2}
\end{equation}

The above equations imply that whenever a novice makes a mistake at
time $t$, the robot takes an action $a_t$ in that state $s_t$ to
positively reinforce the participant, who on the other hand takes an
action $a^\prime_{t}$ and rectifies the mistake moving to the next
state. Here the state is the pattern-making task in the case of
Baxter, while in the case of Tetris, the players are asked to restart
the game rectifying the mistake. Hence the reinforcement learning
agents give rise to a sequence or trajectory that looks like the
following if the novice keeps on making a mistake: $S_t, A_t(R_t)
,A^\prime{_t} ( R^\prime{_t}),S_{t+1},A_{t+1} (R_{t+1})
,A^\prime{_{t+1}} (R^\prime{_{t+1}}),S_{t+2}$. The above sequence
denotes the condition if a participant keeps on making a mistake at
one point. The sequence will stop with the correct action of the
novice learner.  However, the robot keeps on evaluating the other
sections and if a mistake occurs again, the same behavior is
repeated. Therefore the state transition probability $T$:


\begin{equation}
p(s^\prime | s, a) \doteq Pr({S_{t+1} = s^\prime | S_t = s, A_{t} = a, A^\prime{_t} = a }) \label{eq:3}
\end{equation}

Robotic mutual reinforcement is based on psychological principles of
social reinforcement and inclusion and is intended to improve skill
transfer by adapting to the reward value systems of an individual.  In
order to effectively teach a skill, the instructor relies on the
principles of learning theory and basic operant conditioning and
positive reinforcement.  Reinforcement, whether through the addition
of a reward (positive reinforcement) or the removal of something
aversive (negative reinforcement) refers to techniques used by
trainers and instructors whose goal is to increase the likelihood of
the behavior being repeated.  Behavioral psychologists rely on the
principles of positive reinforcement as the primary means through
which to teach and shape behaviors in both humans and animals.  This
type of shaping is considered ideal since the individual participant
or subject is rewarded for the correct behavior and associates a
specific behavior with a specific reward.  The subject is therefore
more likely to repeat the desired behavior in the future (as opposed
to negative reinforcement when the subject is trialing behaviors to
avoid an aversive stimulus).  In the case of mutual reinforcement, the
robot engages in activities and behaviors that positively reinforce
the correct behavior of the participant.  Learning theory dictates,
however, that the value of rewards in positive reinforcement are
subjective in nature and highly dependent upon the individual.  This
means that a reward ($R$) may be of high value to one individual and
of no value to another.  For adequate learning to occur, the reward
must therefore be tailored to each individual.  In the case of mutual
reinforcement, the robot is equipped with a range of rewards that
might be of value to the subject.  Given that all the subjects are
human, the programmed rewards are tailored to humans and were designed
based on the species, culture, and potential individual differences of
the target population.  Humans are social and, in performance
environments, are highly responsive to social inclusion and exclusion
\cite{cheung2015way}, making social signals of high value to almost
every human. The social rewards are designed to mimic
culturally-appropriate interactions to which the participants are
accustomed, which promotes comfort and relaxation and builds
rapport. The ability to adapt and read the human through behavioral
feedback establishes a baseline of communication and language that is
novel and unique between the robot and the human, effectively
mimicking normal social relationships and social learning paradigms.
When a autonomous machine is equipped with various means of social
reinforcement in combination with algorithms that allow for
adaptations to individuals, there is a greater chance of finding a
reward that is of high subjective value for each participant.  This
can only be done if the machine is first equipped with these tools and
then programmed in a way that allows it to perceive, adapt, and adjust
rewards based on the feedback from participant progress.  Mutual
reinforcement is therefore a promising approach to skill transfer
between a robot and human.

\begin{algorithm}
\SetAlgoLined
\KwIn{$\mathit{V_n} \gets \{\mathit{set\ of\ items\ with\ uniform\ weight\ distribution}\}$}
\KwOut{$\mathit{V_{n_{updated}}} \gets \{\mathit{set\ of\ items\ with\ updated\ weights}\}$}

${V_{n_{i^{th}}}}$ ${is\ given\ out\ on\ the\ on\ the\ basis\ of\ weighted\ random\ selection}$\\
\While{${V_{n_{updated}}}, (mistakes) $}{
    \eIf{${V_{n_{i^{th}}}} \gets success$}{
      ${V_{n_{i^{th}}}} = {V_{n_{i^{th}}}} + \alpha$\;
      ${V_{n-{n_{i^{th}}}}} = {V_{n-{n_{i^{th}}}}} - (n-1)/{\alpha}$\;
    }{
      ${V_{n_{i^{th}}}} = {V_{n_{i^{th}}}} - \alpha$\;	      
      ${V_{n-{n_{i^{th}}}}} = {V_{n-{n_{i^{th}}}}} + (n-1)/{\alpha}$;}
        {$EWMA(V_n) = (V_{n_{updated}} * \phi)  +  (Previous\ EWMA(V_n) * (1 - \phi ))$}
	{$V_{n_{updated}} = EWMA(V_n) + \sigma (V_n) $} \\
	$V_{n} = V_{n_{updated}}/{(V_{n_{updated}}.sum)}$ 
    }
\caption{Mutual Reinforcement Learning}
\label{alg:one}
\end{algorithm}


The following sections illustrate Algorithm \ref{alg:one} and the
notations associated with it. It also explains the choice of selection
of certain parameters and their impact in the experiment.

\subsection{Optimization of Reinforcers}

The above algorithm is implemented in both robot and computer gaming
platforms inducing MRL. This method is directly applied to the problem
of searching for an appropriate reward channel preferred by individual
human participants during the skill transfer task. Here each
reinforcer is influenced by the participant's cognition and
performance, and this evaluation directs which reinforcer will be
considered next. Hence, in this method, the robot and the human are
successively generating and evaluating attempts to obtain incremental
improvements for each other. Fig \ref{fig:one} refers to the MRL
concept where the participant rectifies the mistake after getting a
positive reinforcer.

In the particular studies reported here, Baxter has four distinct
reinforcers at its disposal, whereas for Tetris there are seven. $V_n$
denotes the weight vector assigned to $n$ reinforcers, which are
initially a uniform probability distribution summing to 1.  The
reinforcers are given out on the basis of weighted random selection to
meet the exploration-exploitation criteria, but since all the
reinforcers are uniformly weighted at the beginning, weighted
reinforcement in the first step is of no significance. If the
reinforcer $V_{n_{i^{th}}}$ given out from the set is a success then
$V_{n_{i^{th}}} + \alpha$. Here $i$ represents the particular
reinforcer that is provided by the robot or the machine to motivate
people.  $\alpha$ is a small positive fraction called the step-size
parameter, which influences the rate of learning. The value of
$\alpha$ is chosen empirically based on the observed performance of
exploration and exploitation. Figure \ref{fig:thirteen} denotes the entropy fluctuation of the system with the suitable alpha
value.  Selection criteria for the value of $\alpha$ are discussed in
detail in the next subsection.  The value of $\alpha$ adds to the
present weight of the reinforcer denoting its success. $(n-1)$ denotes
the number of remaining reinforcement strategies, and the value of
$\alpha$ is equally distributed among them. If ${V_{n_{i^{th}}}}$ is
successful, $\alpha$ is added to it and $\frac{(n-1)}{\alpha}$ is
subtracted from all the rest of the reinforcers.  In contrast, if
${V_{n_{i^{th}}}}$ is a failure, then $\alpha$ is subtracted from it
and $\frac{(n-1)}{\alpha}$ is added to the rest of the reinforcers.

After this step the weights of $V_n$ are updated.  The mutual
information shared among them is obtained from several interactions
and the values of the reinforcers get updated every time with
probabilities associated with faster skill transfer. The robot tries
to learn about the person's behavior and performance level and then
applies this knowledge to motivate the individual.  We used an
exponential weighted moving average (EWMA) to gain information about
the most recent interactions. For Baxter, since the number of
reinforcers is fewer, we used the past three interactions and for
Tetris we used five.  EWMA only provides information about recent
interactions, but we need to understand the variability of the
reinforcers' success over a longer term. Hence we maintain the value
of two standard deviations $\sigma(V_n)$ over the EMWA values in order
to better notice and interpret success.  Then again the probabilities
of all the reinforcers in the set are updated and prepared for the
next interaction. The robot stops giving out the reinforcers when the
participant stops making mistakes.

MRL, an implementation of which we have described in the above
algorithm, is a novel concept in the field of traditional
reinforcement learning and can be implemented in several algorithmic
approaches to get significant results.

\subsection{Choice of Parameters}

In Algorithm \ref{alg:one} we used several parameters whose values are
tailored depending on the experimental requirements. The parameter
$\alpha$ is chosen on an empirical basis. We used numbers ranging from
0.01 to 0.07 in case of Tetris and 0.01 to 0.04 in case of Baxter. The
range of numbers are selected on the basis of the number of
reinforcers used in the experiment. We calculated entropy $\hat{H}$
depending on the weight of $\alpha$ to determine the mutual
information gain over interactions and whether the autonomous agent is
optimally exploring and exploiting. Using the maximum entropy
principle, we know \cite{shannon1948mathematical} that entropy reduces
over time with the information gain. We calculated the entropy using
values chosen from the ranges given above to determine the most
suitable $\alpha$ value.

\begin{figure}
  \begin{subfigure}[(a)]{0.5\textwidth}
    \includegraphics[width=7cm,height=5cm]{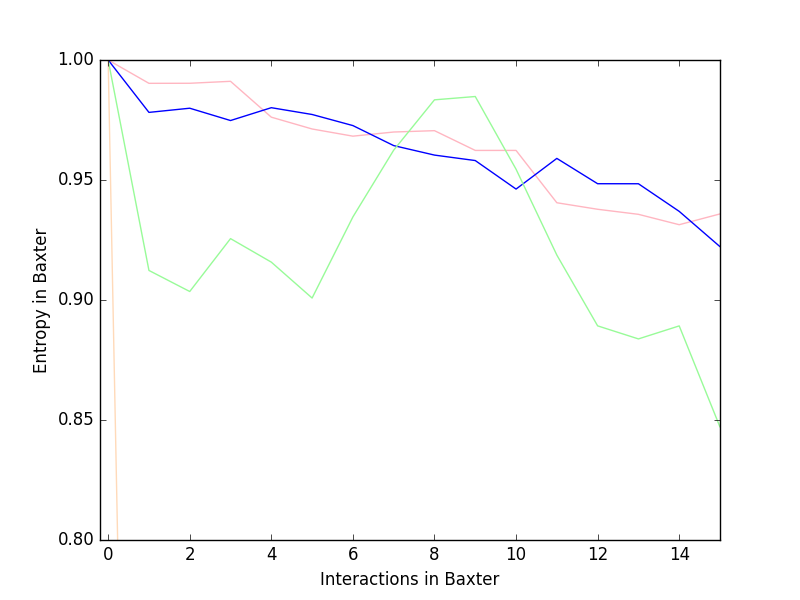}
  \end{subfigure}
  \begin{subfigure}[(b)]{0.5\textwidth}
    \includegraphics[width=7cm,height=5cm]{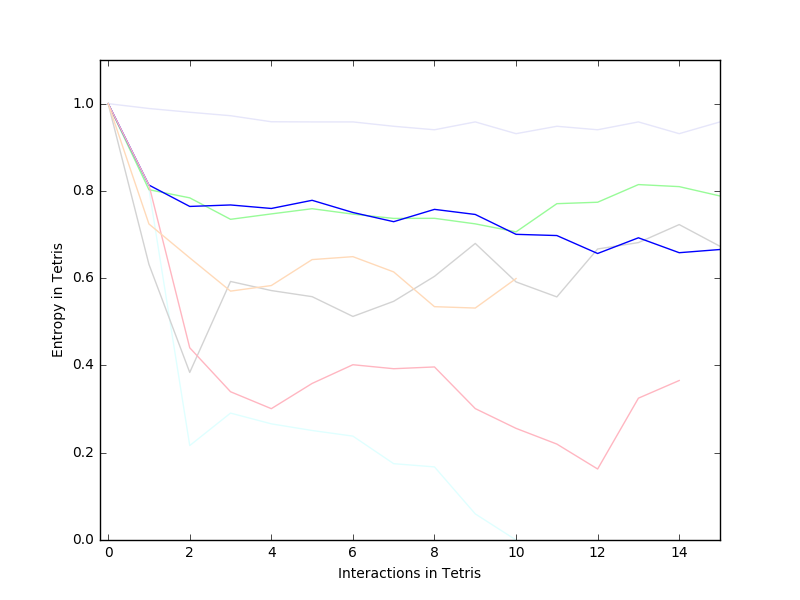}
  \end{subfigure}
  \caption{The figure shows best $\alpha$ values of (a)Baxter and
    (b)Tetris.  $\alpha$ values are chosen on the basis of entropy and
    the exploration exploitation trade-off. The blue line is
    decreasing nearly monotonically, showing steady information gain
    over time.}
  \label{fig:thirteen}
\end{figure}

In Figure \ref{fig:thirteen} we show various rates of
entropy decrease for different $\alpha$ values. The lines in lighter
color show different $\alpha$ values we considered for Baxter and
Tetris and and the dark blue line exhibits the one we selected for our
experiments, namely 0.015 for Baxter and 0.05 for Tetris. These values
not only gain information linearly over time but also trade off
satisfactorily between exploration and exploitation. In the process of
information gain, the rate of decrease of entropy is not always
perfectly linear; all the graphs of different $\alpha$ values are
accompanied by spikes due to exploration-exploitation tradeoffs. We
designed the above algorithm not to behave greedily because exploring
different reinforcers is necessary while breaking the monotony of the
task.  From these figures we can see that at the beginning of the
experiment the entropy is maximum for all $\alpha$ values and
gradually decreases over interactions.

The value of $\phi$ is selected in such a fashion that the robot
considers the last three interactions of Baxter and five for
Tetris. $\phi$ is denoted as the multiplier. In our previous work
\cite{roy2018reinforcement}, we used the robot's experience from the
beginning to the task for the reinforcer selection and found out that
sometimes people prefer more than one reinforcer.  We theorized that
their preference might have changed over interactions, and hence
focused on recent interactions for better performance.

\subsection{MRL and cognitive models}

In traditional reinforcement learning designing an appropriate reward
signal is a critical part of the application process. Various
researchers have coined novel techniques to solve this issue
\cite{Abbeel-ICML-04}.  In contrast, processes like inverse
reinforcement learning (IRL) learn from a expert's behavior, where an
agent tries to infer the reward signal to achieve a particular goal.
In neither of these cases does any two-way interaction between the
expert and the agent take place, and therefore they do not gain the
advantage of situational feedback which is important during a learning
process. To achieve a particular task, both expert and novice both
should exchange feedback through appropriate reward channels. In
practice, designing appropriate reward signals is often an informal
trial-and-error search for a reward signal that produces acceptable
results. In MRL, the expert explores and exploits the reward signals
in the course of judging the novice's actions and performance while
accomplishing the task. Hence if the novice learns slowly, fails to
lean or learns incorrectly the expert cooperates to improve the
student's learning during the process. This is a sophisticated way to
find good reward signals, since feedback is given while accomplishing
a subgoal and the expert can slowly guide the agent towards the
overall goal. Hence unlike other reinforcement learning strategies,
MRL is a complete model that supports task learning with human-robot
interaction simultaneously learning about the reward preferences. In
MRL, since the expert cooperates with the novice during the learning
process, it also becomes aware of the cognitive models involved, which
in turn leads to the design of better reward signals.  To explore the
efficiency of the process, we calculated the machine's regret and the
mutual information shared between the agents using Shannon's entropy
$\hat{H}$. Regret is defined as the difference between the reinforcer
with maximum weight and the reinforcement strategy selected, i.e. $R =
s_{max} - s^{+}$.

Property 1: \textbf{In MRL, an autonomous agent fails to identify the
  cognitive orientation of a participant if it crosses $k$ time steps as
  after $k$ steps no change in entropy occurs, which means no
  information gain.} 

\[
M=
  \begin{bmatrix}
    \dfrac{1}{N} & \dfrac{1}{N} & \dfrac{1}{N} & \dfrac{1}{N} & \dfrac{1}{N} & \ldots \\
    \dfrac{1}{N}+\alpha & \dfrac{1}{N}-\beta & \dfrac{1}{N}-\beta & \dfrac{1}{N}-\beta & \dfrac{1}{N}-\beta & \ldots \\
    \dfrac{1}{N}+\alpha-\beta & \dfrac{1}{N}-\beta-\beta & \dfrac{1}{N}-\beta-\beta & \dfrac{1}{N}-\beta+\alpha & \dfrac{1}{N}-\beta+\beta & \ldots \\
    . \\
    .  \\
    .   \\
  \end{bmatrix}
\]

The above matrix $M$ is a transition matrix with state space $S$,
where $|{S}|=X$ is possibly infinite. $N$ denotes the number of
reinforcers used and $\alpha$ and $\beta$ are the weights that are
added to the system depending on the exploration-exploitation
tradeoff. The above matrix is prepared on the basis of Algorithm
\ref{alg:one}. Now let $\pi^{T}$ be a row vector denoting a
probability distribution on $S$: so each element $\pi_{i}$ denotes the
probability of being in state $i$, and $\sum_{i=1}^{X} \pi_{i} = 1$,
where $\pi_{i} \geq 0$ for all $i = 1, \ldots, X$. The probability
distribution $\pi^T$ is an equilibrium distribution for the above
matrix if $\pi^T M= \pi^T$. That is, $\pi^T$ is an equilibrium
distribution if

$(\pi^{T} M)_{j}=\sum_{i=1}^{X} \pi_{i}p_{ij} = \pi_{j}$  for all $j = 1 \ldots X$. 

That is, $\pi^{T}$, $\pi^{T+1}$ will have the same values and so
on. This is because the values achieve numeric stability after
$\pi^{T}$. That means $M^{t}$ converges to a fixed matrix with all
rows equal as $t\rightarrow\infty$. At this point, no further change
in the Shannon entropy $\hat{H}$ for $\pi^{T}$, $\pi^{T+1}$ will be
observerd.  Entropy in a system denotes its information gain: a
decrease in entropy means more information gain. Here $\pi_{1}$ has
the maximum entropy 1 which decreases linearly over time with
information gain. Now if $\hat{H}$ does not change with time $t$, that
means the robot is not gaining any further information about the
participant's cognitive orientation. Hence we can conclude that
cognitive orientation of a participant can only be found $\leq k$
steps in the transition matrix.
  
Property 2: \textbf{If MRL converges to stationary distribution over
  time (equilibrium), it is independent of the reinforcers used during
  the interaction.}

From property 1, we determined the convergence criteria of $M$. Hence
we know that matrix $M$ converges to $\pi^{T}$,$\sum_{i=1}^{X} \pi_{i}
= 1$ for a large value of $t$. Now the stationary or equilibrium
distribution can be found out by solving $\pi(M-I)=0$, where I is the
identity matrix. If a matrix M reaches equilibrium at
$t\rightarrow\infty$, we know that the cognitive orientation of the
candidate is undetermined. Hence we assume that when equilibrium is
achieved all the reinforcers$(r_1,r_2,r_3,....)$ are utilized and they
failed to affect the human cognition. Hence we can conclude that if
MRL converges to the equilibrium distribution, then it is independent
of the reinforcers used during the interactions.

\section{Reinforcers}
\label{reinforcers}

The autonomous agents weigh several positive reinforcers in this
research to motivate the students if they commit any mistakes
\cite{roy2019mutual}. In this section we discuss the reinforcers used
by the robot and the computer during the experiment and the
effectiveness of Simpson's psychometric model.

\subsection{Reinforcers used with Baxter}
When mistakes are made, Baxter forms a sad face and gives out the
positive reinforcer to encourage participants \cite{Fitter-ICSR-16},
and when they perform correctly after the correction it forms a
smiling face. Other than that, Baxter maintains a neutral face
throughout the task. The following reinforcers are given out depending
upon the subject's assigned experimental group.
\begin{enumerate}
\item \textbf{Verbal reinforcer} : When using this reinforcer, the
  robot asserts that it is trying to encourage the subject with some
  positive feedback. Since Baxter does not have its own audio
  interface, we used speakers to produce the robot voice. In our
  experiment, if the subject makes a mistake, the robot will verbalize
  something like, ``Sorry dear, don't worry. You can do it''.

\item \textbf{Hint-based reinforcer} : This takes the form of a hint
  given to the participant during a task.  The hint does not provide
  the correct answer but tries to influence the subject's thought
  process so that it increases the learning rate of the participant.
  For example, during the pattern making process, if a candidate
  places an incorrect marker, Baxter suggests flipping the marker box
  and trying the other side, before rejecting the block entirely.
  Thus people can track the blocks they have already tried to place in
  a particular spot.

\item \textbf{Simple-feedback reinforcer}: In this case, the robot
  only identifies the correct or the incorrect marker. It doesn't
  attempt to induce any kind of positivity or motivation in the
  participant. This is because some people are not fond of external
  motivations and only a rectification in the task can influence them
  to perform better.

\item \textbf{Gesture-based reinforcer}: In this case, the robot adds
  a consoling gesture by patting at the student's back and also
  provides positive verbal feedback as referenced above.

\end{enumerate}

\subsection {Reinforcers used with Tetris}

In the case of Tetris, seven different positive reinforcers are used
during the interactions.  We increased the number of reinforcers
because in a fast-moving gaming scenario, we anticipated more
interactions per session.  All the positive reinforcers are displayed
in an audio-visual setting whenever a participant makes a mistake.
Here all of the reinforcers provided some sort of hint for the player
to perform better. For example, whenever a player is playing for too
long without scoring any points, reinforcers are provided such as
``Clear the lines quickly for faster score'' or guiding the player to
check for the upcoming blocks to plan the next move ahead. The type of
incentives are manipulated according to the platforms we employed to
demonstrate the effectiveness of MRL.


\subsection{Simpson's psychometric model}
In both of the above platforms, Simpson's psychometric model
\cite{simpson1972classification}\cite{simpson1966classification} is
used to identify the skill transfer. This model characterizes the
principles of skill evaluation, which are closely linked with
important aspects of human cognition. It is widely used by teachers,
professional specialists and scientists to evaluate curricular
problems with greater precision. Simpson's psychometry domain is
defined in form of a taxonomy which gives us a clear idea about how
knowledge is acquired by an individual and how that is later applied
to execute tasks. Simpson's psychometric model is broadly classified
into Perception, Set, Guided Response, Mechanism, Complex Overt
Response, Adaptation and Origination. Perception is related to the
awareness of the present situation.  Set is the eagerness of the human
participant to volunteer for the task. Guided Response is the early
stage of learning a complex skill with the help of an
instructor. After the participant has learned the task, the later
stages of the psychometric domain involve applying the
training. Mechanism is the immediate step to demonstrate basic
proficiency with respect to a simple application. Complex Overt
Response is associated with skillfully applying complex versions of
the same task with greater proficiency. Adaptation signifies complete
learning, where individuals can respond to uncertain events, while
Origination is the last phase of learning where humans can generate
new ideas from their knowledge. Here the experiments are designed to
confirm the feasibility of Simpson-based skill evaluation. Since the
tasks are designed in a lab setting, we only used a few of the above
categories to determine the skill transfer process. The following
section discusses the experimental models and the findings associated
with them.

\begin{figure}
\begin{subfigure}[(a)]{0.45\textwidth}
  \includegraphics[width=5cm]{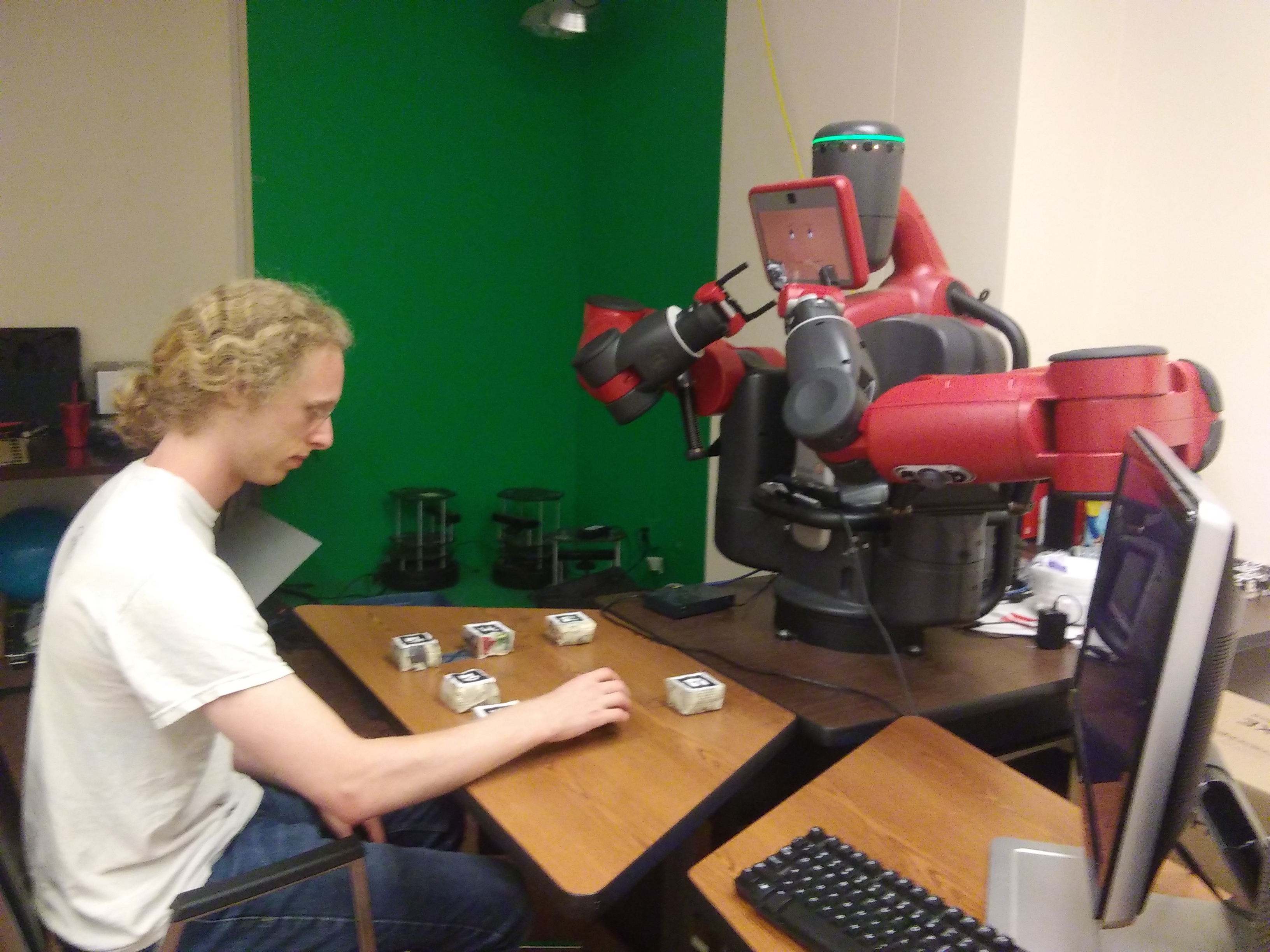}
\end{subfigure}
\begin{subfigure}[(b)]{0.45\textwidth}
  \includegraphics[width=5cm]{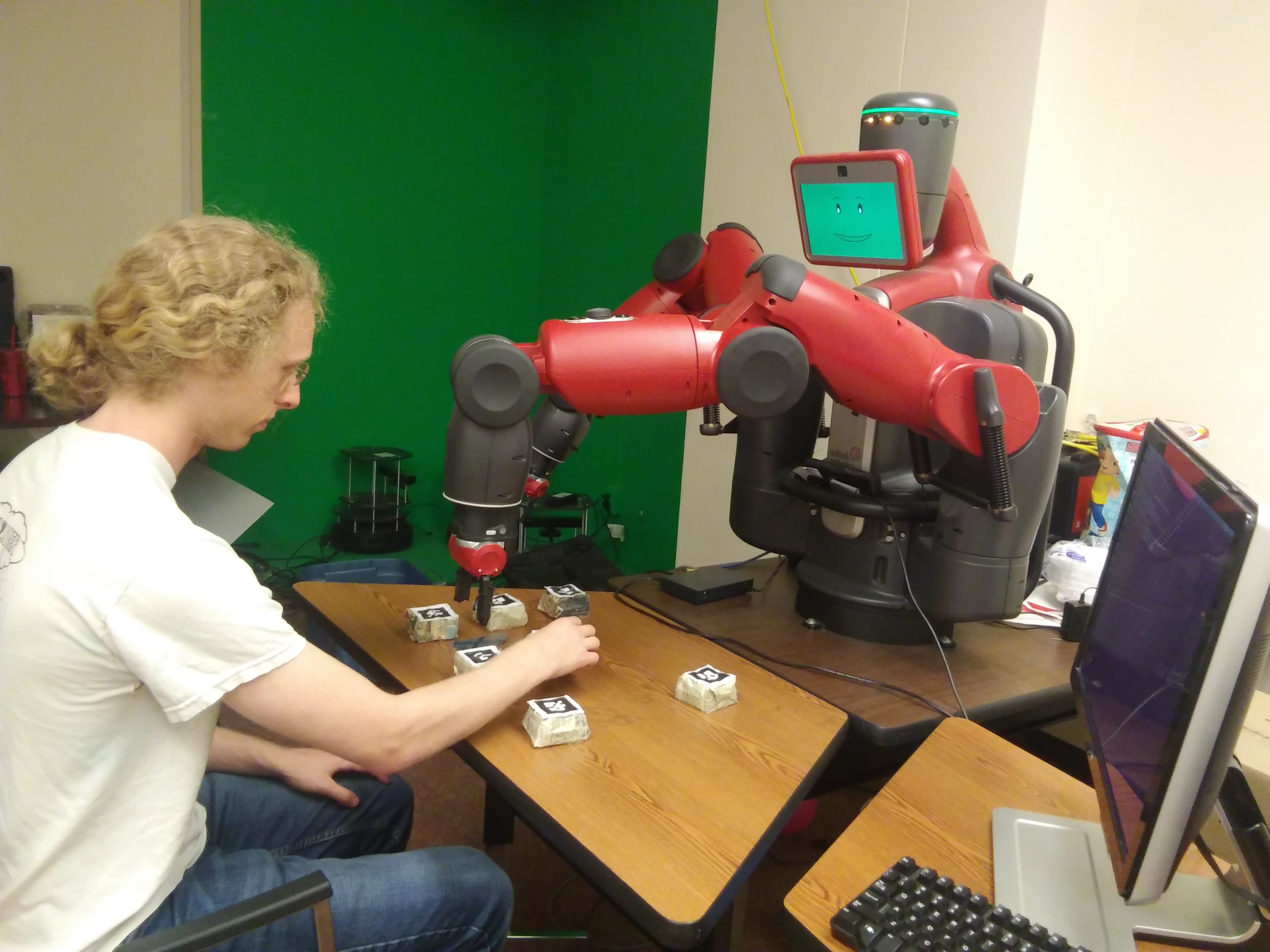}
\end{subfigure}
\begin{subfigure}[(c)]{0.45\textwidth}
  \includegraphics[width=5cm]{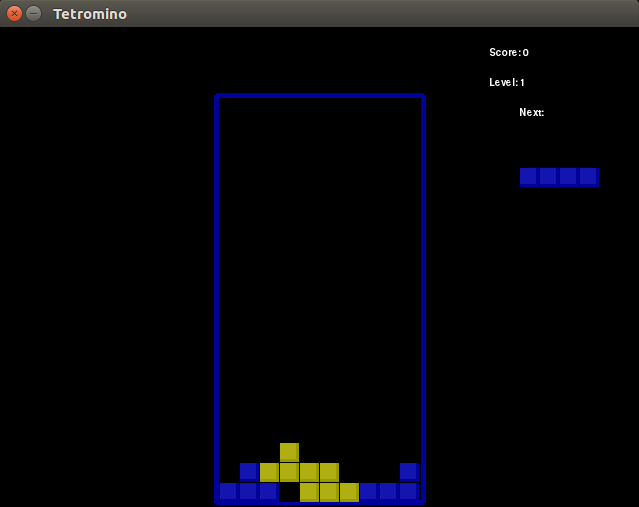}
\end{subfigure}
\begin{subfigure}[(d)]{0.45\textwidth}
  \includegraphics[width=5cm]{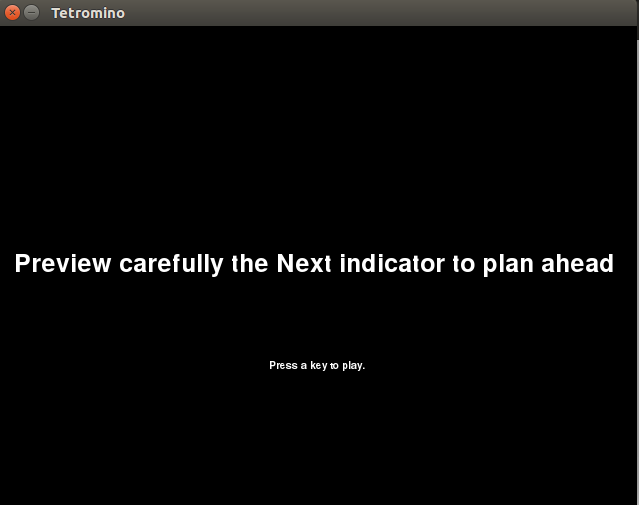}
\end{subfigure}
  \caption{The figure shows the experimental scenario where the
    participant is interacting in an MRL framework with Baxter and
    Tetris.  (a) Participants make a mistake and Baxter forms a sad
    face while providing a positive reinforcer to encourage the
    participant. (b) The mistake is rectified by the participant and
    Baxter forms a happy face. (c) The participant has made a mistake
    in the Tetris game (allowed a gap to form in a line) (d) A
    reinforcer is provided to rectify the mistake.}
  \label{fig:fourteen}
\end{figure}

\section{Task description with Baxter}

$n=34$ (age : $\mu=19.69$, $\sigma=3.47$, male=13, female=21, none=11,
random=11, learned=12) participants were recruited for the experiments
with Baxter, which ran for a time $t$ ($\mu=18.47 $,$\sigma=5.60$ in
minutes). Among the subjects, 75\% had never interacted with a robot,
8.30\% interacted a year ago, and 5.50\% each a month and a week
ago. The task involved in the experiment was divided into two large
sections that was further divided into two smaller subsections. The
tasks are designed to observe successful skill transfer from robot to
human using Simpson's psychometric model, where each category in the
taxonomy transitions to another with the goal of skill transfer. In
this experiment we only used Guided Response, Mechanism and
Adaptation.  Guided Response I and II occur in the first half of the
experiment where the robot first teaches the participant about the
augmented markers and then motivates them throughout the learning
process. During this the robot also evaluates the performance of the
participants. In the second half of the first section the robot
teaches the participant a complex pattern with dual-faced augmented
markers and asks the student to reconstruct it. Again, during this
process, the robot positively reinforces the learner with simple
yes-or-no, random or learned MRL feedback depending upon their
assigned experimental group. Participants are allowed to observe the
pattern making process and then the markers are immediately shuffled
and they are asked to start the reconstruction immediately. Baxter
transitions its left hand camera from one spot to another for
evaluation. Baxter does not progress to the next position until the
participant rectifies any mistakes. During this process the
participants get hint, simple, verbal or gesture feedback depending
upon their group and Baxter with their performance tries to identify
their cognitive orientation. Fig \ref{fig:fourteen} (top frames)
corresponds to the experimental procedure with Baxter.

In the second half of the experiment the participants are asked twice
to reconstruct the pattern, this time without any motivation, to
observe how well the skill transfer succeeded.  In the last half of
the experiment participants are asked again to identify two random
markers from the set they were taught at the beginning of the
experiment to analyze adaptation.  Each participant in the experiment
was assigned different complex patterns for reconstruction. The marker
placement at each position depends upon the robot.  At the end of the
experiment, the subjects were given questionnaires to answer using a
5-point Likert scale.

\subsection{Results}

The results section is further divided into subsections discussing the
performance of the participants, the mutual information shared, the
robot's regret and the mental model of the participants during the
task.

\subsection{Subjective performance evaluation}

\begin{figure}
\begin{subfigure}[(a)]{0.5\textwidth}
  \includegraphics[width=6cm,height=6cm]{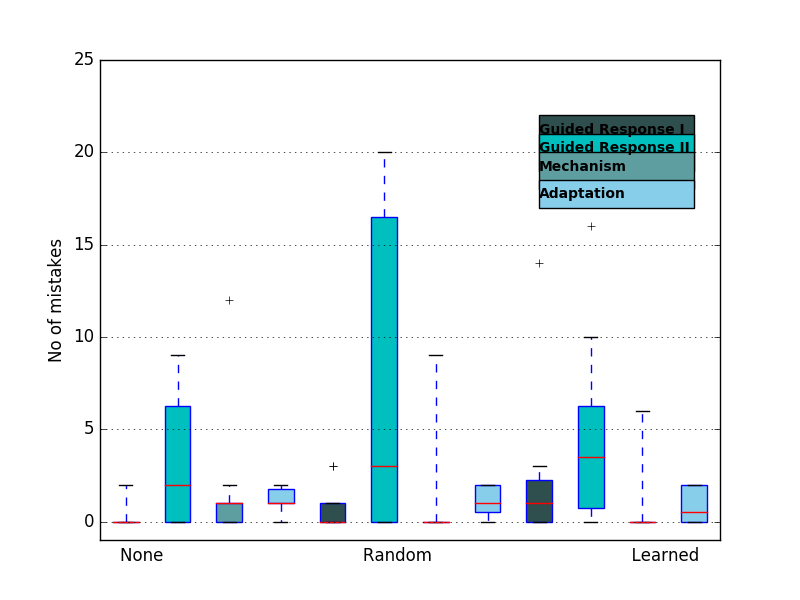}
  \end{subfigure}
\begin{subfigure}[(b)]{0.5\textwidth}
  \includegraphics[width=6cm,height=6cm]{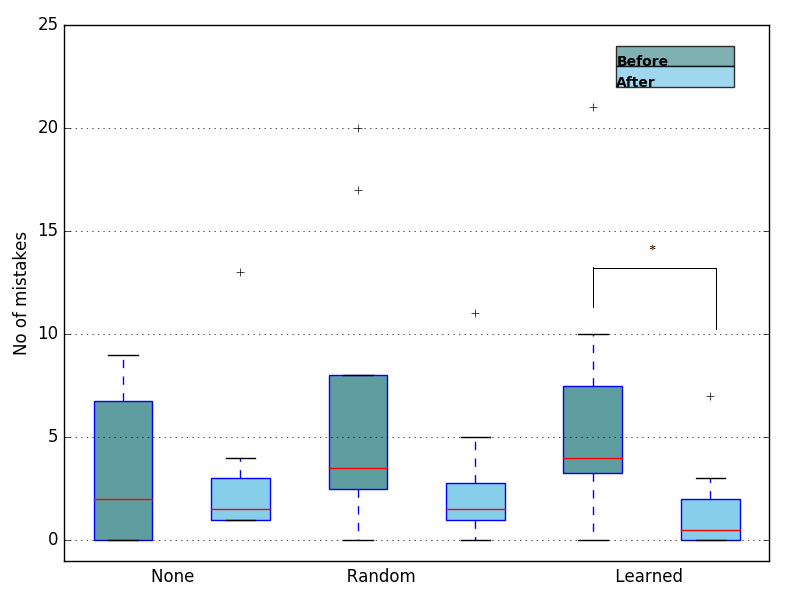}
  \end{subfigure}
  \caption{(a) Skill transfer is analyzed different levels of Simpson's psychometric model. (b) Mistakes made by the participants before and after the skill transfer while interacting with the Baxter. MRL improves skill transfer ($p < 0.05$)}
  \label{fig:four}
\end{figure}

To quantify the skill transfer procedure, we calculated the number of
mistakes made by each participant in all of the groups in different
phases of the experiment. Figure \ref{fig:four} shows the number of
mistakes according to Simpson's psychometric model. There are
participants ($\approx 8.82\%$) who didn't commit any mistake
throughout the experiment, so their results are not included in the
mistakes data. From Table \ref{table:tab1} and Figure \ref{fig:four}
we can see that the number of mistakes made during the Mechanism and
Adaptation phases are significantly less than during the Guided
Response phases irrespective of the groups, which shows the effect of
robot feedback during the task. Again if we compare the skill transfer
among the groups in the figure, we can see that the number of mistakes
in the learned MRL group is comparatively lower than the other
two. Table \ref{table:tab1} presents the performance of the
participants in all the phases of the experiment.  Figure
\ref{fig:four} also plots the number of mistakes in each group before
(Guided Response I and II) and after (Mechanism and Adaptation) the
skill transfer procedure; we see that the participants made
comparatively fewer mistakes in the learned group than in the other
two. Using a t-test for skill transfer outcomes while comparing the
learned group with random reinforcers we get a significant $p$ value
$<$ 0.05. Thus we can conclude that MRL has successfully worked in
terms of the skill transfer procedure.

\begin{table}[htbp]
  \caption{The table below shows the mean and standard deviation
    values of the performance of the participants in different phases,
    viz. Guided Response I , Guided Response II, Mechanism and
    Adaptation using Simpson's psychometric model in the skill
    transfer scenario in Baxter.}
\begin{center}
\begin{tabular}{|c|c|c|c|c|c|c|}
\hline
\textbf{Group} & \multicolumn{2}{|c|}{\textbf{None }} & \multicolumn{2}{c|}{\textbf{Random}} & \multicolumn{2}{c|}{\textbf{Learned}}  \\
\cline{2-7}
\hline
& \textbf{\textit{M}} & \textbf{SD} & \textbf{\textit{M}} & \textbf{SD}  & \textbf{\textit{M}} & \textbf{SD}\\
\hline
Guided Response I  & 0.4 & 0.84 & 0.8 & 1.17 & 2.08 & 3.92  \\
\hline
Guided Response II & 2.08 & 3.92 & 7.45 & 8.37 & 4.5 & 4.80 \\
\hline
Mechanism & \textbf{1.8} & 3.65 & \textbf{1.33} & 3.04 & \textbf{0.64} & 1.80 \\
\hline
Adaptation & \textbf{1.2} & 0.63 & \textbf{1.18} & 0.87 & \textbf{0.83} & 0.94 \\
\hline
\end{tabular}
\label{table:tab1}
\end{center}
\end{table}

\subsection{Entropy analysis}

\begin{figure}
  \includegraphics[width=6cm,height=6cm]{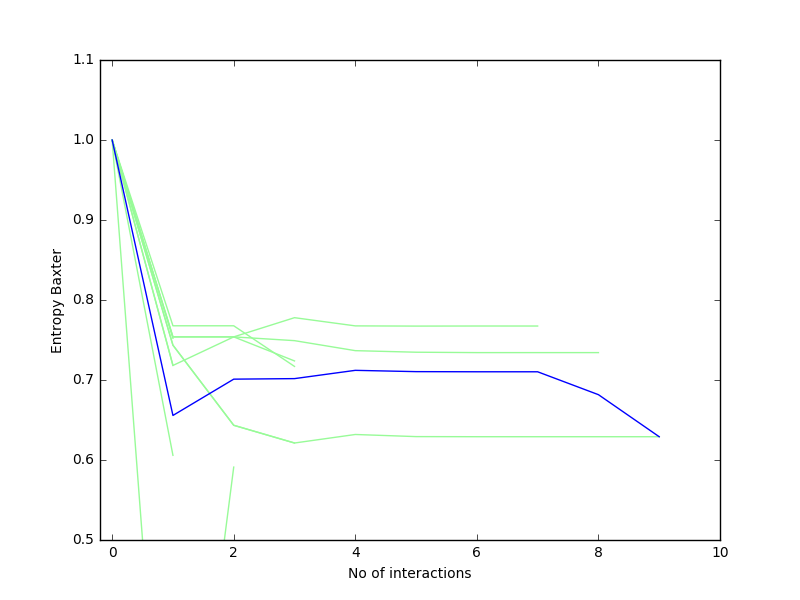}
  \caption{Mean entropy (blue) of the information of the robot
    collected from the interactions with the participants (green).}
  \label{fig:five}
\end{figure}

Entropy denotes the randomness of a system. In Figure \ref{fig:five},
we can see that the entropy of the information of the robot obtained
by interactions goes down monotonically along with each interaction
with its human participant in MRL.  With each interaction, the robot
is gaining more information about the participants' performance and
their cognitive orientation towards each reinforcer. The pale green
lines show the entropy of the information of the each participant
obtained by interactions with Baxter, and the blue line is their mean
performance.  Not every participant had an equal number of
interactions with Baxter, but regardless, using Algorithm
\ref{alg:one}, the machine manages to gain information steadily about
their performance. For each participant, the value of entropy varied
depending on the exact pattern of the robot's choices of exploration
and exploitation, but we can conclude that for each participant it has
gained some information at each interaction which later helped it to
construct a successful mental model.

\subsection{Regret analysis}

\begin{figure}
  \includegraphics[width=6cm,height=6cm]{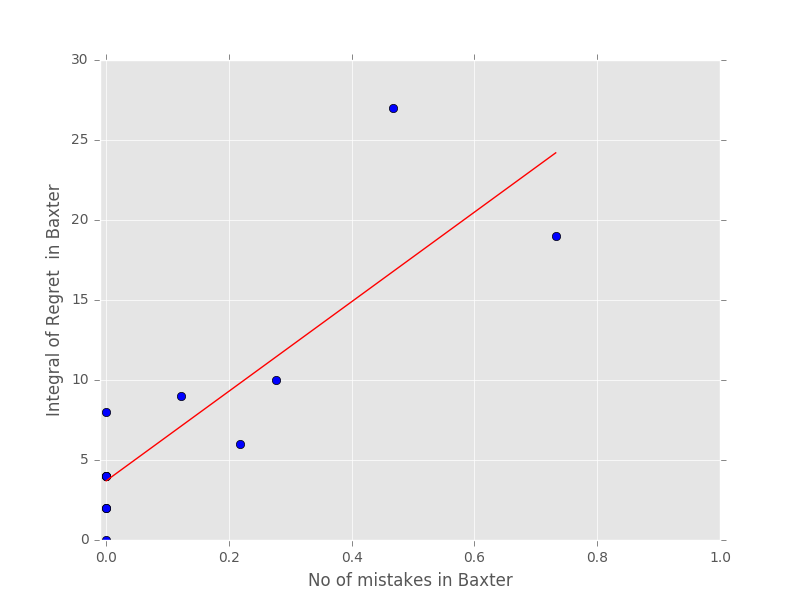}
  \caption{Regret analysis of Baxter. The linear curve shows the best
    fit with coefficient $r = 0.85$.}
  \label{fig:seven}
\end{figure}

As mentioned in Section 3.3, regret is calibrated on the basis of the
decision making ability of the robot. It depends upon the subject's
performance, which helps in characterizing the most appropriate
reinforcement learning strategy. We correlated the number of mistakes
made by the participants and the total regret felt by the robot
(Figure \ref{fig:seven}) and found a linear relationship between the
number of mistakes and the robot's regret. The value of the
coefficient is $r=0.85$; thus the robot's regret is strongly
correlated and the reinforcement learning strategy used by the robot
to understand human responses and improve their performance, is
working appropriately. For the participants who had several
interactions with the robot or made many mistakes, Baxter tried to
explore different reinforcement strategies at different times, trying
to increase their learning rate.  This illustrates that Baxter can
successfully train people to achieve complex task using their
preferred motivations.

\subsection{Mental model analysis}

At the end of the experiment, those participants in the learned model
group were asked to choose their preferred reinforcers.  Baxter could
correctly identify the preferred reinforcers in half of the cases
(twice as effectively as a random baseline).  Thus, MRL allowed the
robot successfully to identify the cognitive orientation of the
participants to a large extent (accuracy score $=$ 0.50). During the
task, since the number of interactions was limited, the robot did not
have sufficient opportunity to engage in the exploitation aspect of
the reinforcement learning, and thus its ability to identify preferred
reinforcers was limited (but still reasonably successful).  In these
experiments, Baxter explored more than exploited, which impacted the
types of reinforcers given out by the robot.

At the end of each experiment we probed participants with a 5-point
Likert scale (1: Strongly disagree; 2: Disagree; 3: Neutral; 4: Agree;
5: Strongly agree). Participants in the no reinforcer group
($\mu$=3.54 , $\sigma$=0.68 ), random reinforcer group ($\mu$=4 ,
$\sigma$=0.77 ), learned reinforcer group ($\mu$=3.25 , $\sigma$=1.22
) wanted to play with Baxter again and again in no reinforcer group
($\mu$=4.0 , $\sigma$=0.63), random reinforcer group ($\mu$=3.63 ,
$\sigma$=0.92) and learned reinforcer group ($\mu$=2.6 ,
$\sigma$=0.89) thought it is useful as a teacher.  Interestingly, the
people who performed poorly during the experiment neither found Baxter
to be useful nor thought it was a good teacher.  In other words,
people who enjoyed the interaction also found it helpful and wanted to
come again to learn from the robot, whereas those were not fond of the
robot were not interested in the experiment and ended up performing
poorly.


\section{Task Description with Tetris}

$n=31$ (age $\mu=19.77$, $\sigma=5.07$, male=9, female=22, none=11,
random=10, learned=10) participants were recruited for the experiments
with Tetris for 15 minutes.  We conducted an experiment in a gaming
scenario to observe the performance of MRL across different
platforms. Among all the participants, $74.19 \%$ of the subjects had
played the game but not within the last year, $12.19 \%$ had played
within the last year but not the last month, and the rest had played
the game within the last month.  Like with the Baxter scenario,
participants were also being trained to be better Tetris players. The
skill transfer scenario is also analyzed with Simpson's psychometric
model (Guided Response and Adaptation).  As is common in Tetris games,
during each move, the next block is shown alongside the 10x20 game
board so that players plan their move ahead. As in the previous
experiment, the teaching process is divided into two phases. Initially
the participants are asked to play the game for 15 mins with
reinforcers provided depending upon their assigned experimental group
(simple, random, MRL).

Whenever the participant makes a mistake, the machine alerts them,
provides a reinforcer, and then they are allowed to continue.  For
Tetris, mistakes are considered to be placing a block in such a way to
hinder fast scoring.  Wrong placement is associated with forming a gap
between lines which will make it difficult to eliminate the line of
blocks in the future.  Also, if a player places blocks several times
in a row without eliminating any lines, that is also considered as a
mistake as the towers of blocks build up and get closer to ending the
game.  During the first experiment task, if the game is lost,
participants are allowed to restart, so that all subjects received the
same time to learn properly. In the second portion of the experiment
(Adaptation), participants are asked again to play the game, this time
without reinforcement training, to assess the quality of their
learning.  They are asked to play until losing, or until 5 minutes
elapsed ($\mu$=2.68 $\sigma$=0.88 minutes).  Since the experiment had
only two phases, we used Guided Response and Adaptation to analyze the
skill transfer.  After the game, the subjects were also given
questionnaires probed with a 5-point Likert scale.

\begin{figure}
  \includegraphics[width=7cm,height=7cm]{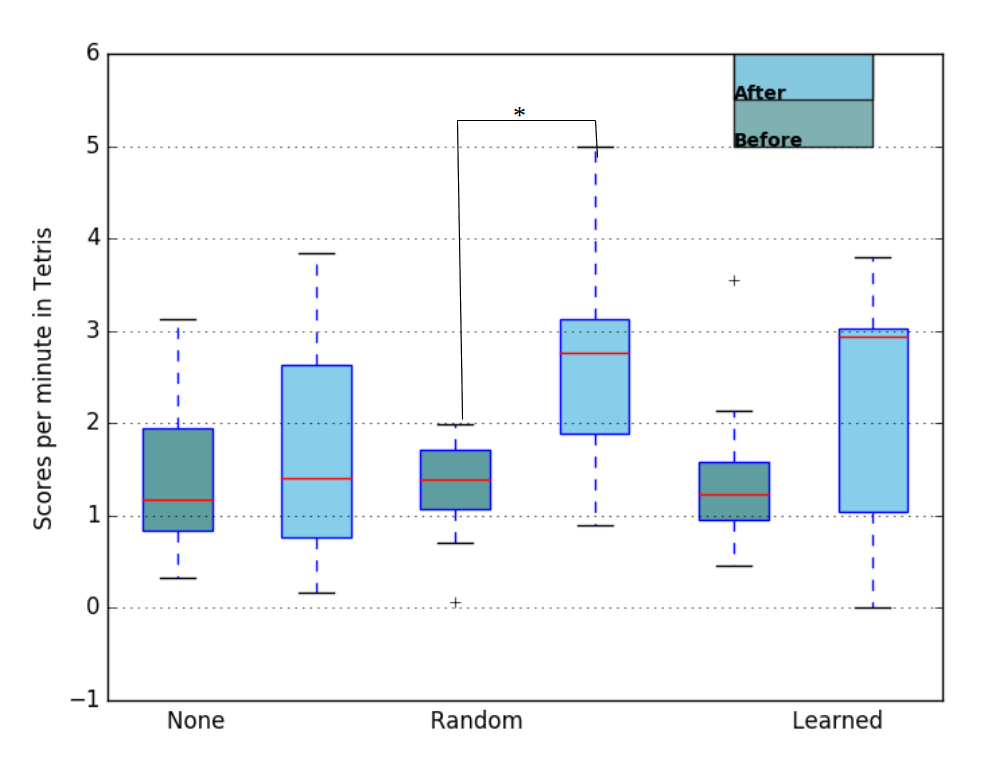}
  \caption{Scores acquired by participants per minute before and after
    skill transfer during Tetris. In the random group, subjects showed
    significant progress after skill transfer p$<$0.05.}
  \label{fig:nine}
\end{figure}

\subsection{Results}

Similarly to the Baxter experiment, our analysis of the Tetris
scenario is divided into subjective and entropy evaluation, regret and
mental model analysis.

\subsection{Subjective performance evaluation}

For Tetris, we computed the skill transfer on the basis of Simpson's
psychometric model. Since, in a game like Tetris, people are expected
to make a large number of mistakes, we computed the scores per minute
of the participants of different groups. Figure \ref{fig:nine} denotes
the scores of the participants before (Guided Response) and after
(Adaptation) the skill transfer procedure.  The subjects in the
random group scored significantly better than the other two
($p<0.05$). Here in case of the random models the participants got some sort of feedback which reinforced them to perform better at the task. The random group showed significant improvement than the none group ( p$\approx$0.01). Hence although the participants responded to feedbacks , there is not much difference between the random and the learned models. Since in this task may people have already played tetris before the MRL didnot have salutary effect on the participants. 

\begin{figure}
  \includegraphics[width=5cm,height=5cm]{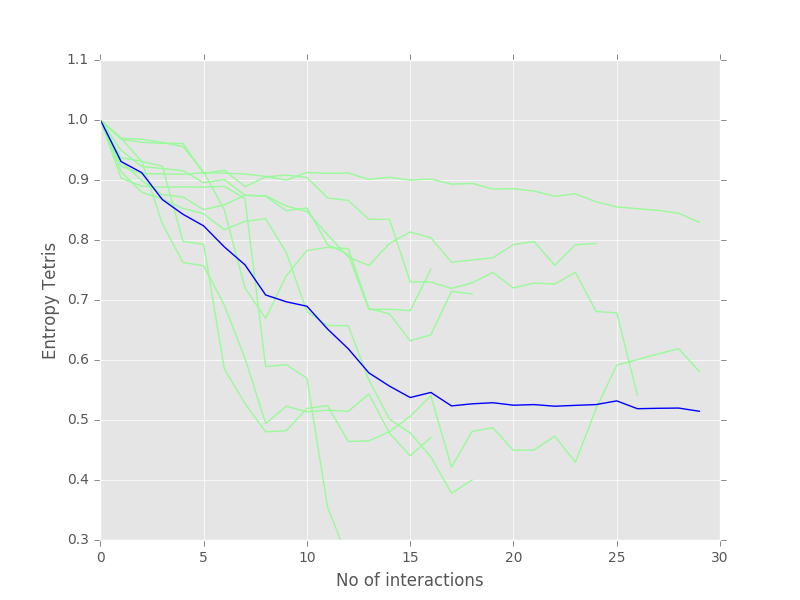}
  \caption{The above asymptotic graph shows the mean entropy (blue) of
    the information of the game system collected from the interactions
    with the participants (green) in Tetris.}
  \label{fig:six}
\end{figure}

\subsection{Entropy analysis} 

As shown in Figure \ref{fig:six}, the entropy of the gaming device's
information monotonically decreases with each interaction with its
human participant in MRL. Like Baxter, this shows that the computer is
gaining information about the participant's performance and its
cognitive orientation towards each reinforcer with each
interaction. The pale green lines show the entropy of each participant
over each interaction with the machine and the blue line shows the
mean performance.

\begin{figure}
  \includegraphics[width=5cm,height=5cm]{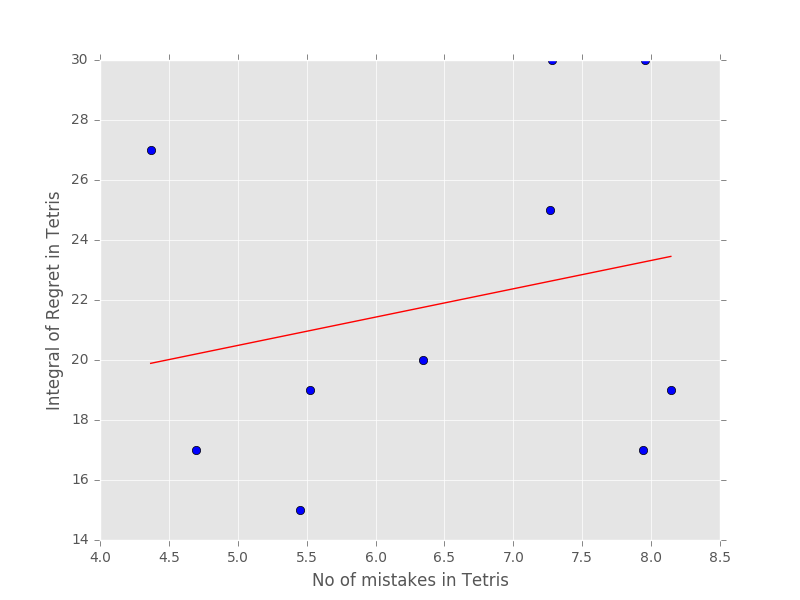}
  \caption{Regret analysis of Tetris. The linear curve shows the best fit with coefficient $r < 0.50$}
  \label{fig:eight}
\end{figure}

\subsection{Regret analysis}

Like Baxter, here also we tried to calibrate the regret of the gaming
system. The correlation coefficient in this case is $r<0.50$. Although
the system tried to explore different reinforcement strategies at
different times, trying to increase the learning rate, the results are
suggestive and not conclusive. Fig. \ref{fig:eight} is the best fit
curve to analyze the relation between regret of the system and the
integral of the number of mistakes made by the participants. This is
because the participants utilized all the reinforcers given to them
without distinguishing between them very much.


\subsection{Mental model analysis}

At the end of the experiment, those participants in the learned model
were asked to choose their preferred reinforcers.  The machine could
partially identify the preferred reinforcers in Tetris.  For example,
reinforcer a is often misclassified as e in many cases.  In Tetris,
since all the reinforcers were fairly similar to one another, in spite
of the fact that they guided participants differently, the subjects
often failed to distinguish the efficacy of one versus
another(accuracy score $<$ 0.50). Hence they were able to make use of
all of the provided reinforcers successfully, without much
differentiation between them.  It was difficult for the system to
determine the preferred reinforcer of the participant, and only it
occasionally successfully identified the best reinforcers. Hence there is not much difference between the performance of the learned group with that of the random group of participants. 

At the end of each experiment we probed participants with a 5-point
Likert scale (1: Not all helpful; 2: Moderately helpful; 3: Neutral;
4: Helpful; 5: Extremely helpful).  The subjects wanted to play Tetris
again with ($\mu$=3.0, $\sigma$= 0.70) none, ($\mu$=3.2, $\sigma$=
0.78) with random and ($\mu$=3.3, $\sigma$= 0.82) with MRL.  They
thought it was useful ($\mu$=2.1, $\sigma$= 0.75) with none,
($\mu$=2.5, $\sigma$= 0.85) with random and ($\mu$= 2.5,
$\sigma$=0.97) with MRL.  Overall, more than half the subjects found
the reinforcers useful and therefore thought the system was a good
teacher.  These statistics show significant results.

\subsection{Discussion}

In the experiment, the results are sometimes not as strong as we might
hope for several reasons.  If the robot's grippers were closed, they
occasionally hindered the camera, blocking the robot from identifying
the markers, since we used the left hand camera for detection and
evaluation purposes.  The Baxter arm and gripper are not extremely
dextrous; it is sometimes very unsophisticated in its attempts to pick
up the blocks whenever they not lying perpendicularly to the camera.
The markers, after several tasks and the degradation which resulted
from repeated handling by both human and robot, became unclear and
difficult for the marker tracking algorithm to recognize, which also
contributed to system crashes.  Again, many of the young adults who
participated in the experiment failed to connect to the robot
emotionally and lacked engagement. Some subjects paid very little
attention to the robot's attempts to communicate a reinforcement
strategy, to the point that the subjects attempted to interact with
the researchers conducting the experiment rather than the robot.  Some
participants simply produced iteration after iteration of patterns
until they happened upon the correct one, without paying attention to
the robot gamely attempting to help.  Rather, they simply tried each
block at each position to figure out the right approach. Hence Baxter
on its end was confused in providing the reinforcement strategy.  For
this reason, we see that Baxter was only able to identify a successful
motivational strategy for half the participants in the learned
group.

Another potential point of alienation came from the fact that
Baxter's voice did not issue from the robot itself, but rather a
speaker off to the side (since the robot hardware lacks sound
capability).  The students had to turn to their right side and
interact with a computer console to give Baxter their feedback in form
of yes or no, which is unsophisticated; verbal interaction would have
been a better option.  However, participants in the experiment came
from different national backgrounds and language abilities, so it was
very hard for the robot to understand their pronunciation, and we were
therefore forced to keep the human feedback in that format.

Seven blocks with 14 markers can be placed in many, many ways, but no
participants required nearly that many attempts to figure out the
correct pattern.  Thus, even when they did not directly engage with
the robot's attempts to teach, it still had some impact on them.  We
used the spatial arrangement of markers as our complex task.  Some
people who performed better in all the groups might simply be good at
this style of task and would fail at some other complex task.  The
difference in performance between the groups might be different in
different complex task scenarios.  Our MRL theory applies to the
people who have performed poorly in the task, and therefore received
appropriate motivations. They might be better at a different complex
task, and therefore engage differently with the reinforcement
behavior.  In the case of Tetris, people are well acquainted with the
game, so the reinforcers might only have a little effect on
them. Although the autonomous agents successfully developed a teaching
strategy for only half of the participants, it is enough to suggest
that such feedback does have impact on human behavior and
learning. Furthermore, this approach allows the autonomous agent to
assess its own success and learn to calibrate its own interactions in
ways that lead to successful teaching.

\section{Conclusion and future work}

In this work, we studied the problem of skill transfer from a robot to
human, where the autonomous agent is not only learning about the human
mental model but also trying to adapt its own accordingly. In this
shared environment, the robot is trying to maximize the cumulative
reward by learning about human behavior and simultaneously improving
its own cognitive model. We highlighted mutual information
communicated between the robot and the human, and validated their
interaction in skill transfer using real-time experiments in both
robot and gaming platforms.  The subjective performance, information
gain over time and the confusion matrices give us a conclusive idea
how robots and computer systems can successfully transfer skills from
themselves to humans.

In our future work we would like to implement MRL across different
platforms. Heavy construction equipment like excavators and backhoes
are required to perform complex tasks like digging, truck loading and
ditch crossing, requiring a series of complex manipulations.  Learning
appropriate manipulations for these different situations is a hard
task.  We want to implement MRL in these scenarios where humans can
learn the subtlety of control manipulations with robot assistance. In
addition, we intend to investigate the necessary behavioral changes
required to be adapted by the robots to become better trainers over
time.  We would also like to involve robots in guiding students
towards correct actions and the responses they should develop, keeping
the learners more thoroughly engaged in the task. We want to design
and identify suitable reward channels for individuals to learn a task
proficiently, simultaneously identifying their mental models in a
robot-human interaction framework.
 
\begin{acks}

This work was supported by NSF award \#1527828 (NRI: Collaborative
Goal and Policy Learning from Human Operators of Construction
Co-Robots).

\end{acks}

\bibliographystyle{ACM-Reference-Format}
\bibliography{sample-bibliography}


\begin{thebibliography}{50}


\ifx \showCODEN    \undefined \def \showCODEN     #1{\unskip}     \fi
\ifx \showDOI      \undefined \def \showDOI       #1{#1}\fi
\ifx \showISBNx    \undefined \def \showISBNx     #1{\unskip}     \fi
\ifx \showISBNxiii \undefined \def \showISBNxiii  #1{\unskip}     \fi
\ifx \showISSN     \undefined \def \showISSN      #1{\unskip}     \fi
\ifx \showLCCN     \undefined \def \showLCCN      #1{\unskip}     \fi
\ifx \shownote     \undefined \def \shownote      #1{#1}          \fi
\ifx \showarticletitle \undefined \def \showarticletitle #1{#1}   \fi
\ifx \showURL      \undefined \def \showURL       {\relax}        \fi
\providecommand\bibfield[2]{#2}
\providecommand\bibinfo[2]{#2}
\providecommand\natexlab[1]{#1}
\providecommand\showeprint[2][]{arXiv:#2}

\bibitem[\protect\citeauthoryear{Abbeel and Ng}{Abbeel and Ng}{2004}]%
        {Abbeel-ICML-04}
\bibfield{author}{\bibinfo{person}{Pieter Abbeel} {and}
  \bibinfo{person}{Andrew~Y. Ng}.} \bibinfo{year}{2004}\natexlab{}.
\newblock \showarticletitle{Apprenticeship learning via inverse reinforcement
  learning}. In \bibinfo{booktitle}{\emph{Proceedings of the 21st International
  Conference on Machine Learning (ICML)}}.
\newblock


\bibitem[\protect\citeauthoryear{Audibert, Munos, and Szepesv{\'a}ri}{Audibert
  et~al\mbox{.}}{2009}]%
        {audibert2009exploration}
\bibfield{author}{\bibinfo{person}{Jean-Yves Audibert},
  \bibinfo{person}{R{\'e}mi Munos}, {and} \bibinfo{person}{Csaba
  Szepesv{\'a}ri}.} \bibinfo{year}{2009}\natexlab{}.
\newblock \showarticletitle{Exploration--exploitation tradeoff using variance
  estimates in multi-armed bandits}.
\newblock \bibinfo{journal}{\emph{Theoretical Computer Science}}
  \bibinfo{volume}{410}, \bibinfo{number}{19} (\bibinfo{year}{2009}),
  \bibinfo{pages}{1876--1902}.
\newblock


\bibitem[\protect\citeauthoryear{Bickhard}{Bickhard}{2009}]%
        {bickhard2009interactivism}
\bibfield{author}{\bibinfo{person}{Mark~H Bickhard}.}
  \bibinfo{year}{2009}\natexlab{}.
\newblock \showarticletitle{Interactivism: a manifesto}.
\newblock \bibinfo{journal}{\emph{New Ideas in Psychology}}
  \bibinfo{volume}{27}, \bibinfo{number}{1} (\bibinfo{year}{2009}),
  \bibinfo{pages}{85--95}.
\newblock


\bibitem[\protect\citeauthoryear{Boccanfuso, Barney, Foster, Ahn, Chawarska,
  Scassellati, and Shic}{Boccanfuso et~al\mbox{.}}{2016}]%
        {boccanfuso2016emotional}
\bibfield{author}{\bibinfo{person}{Laura Boccanfuso}, \bibinfo{person}{Erin
  Barney}, \bibinfo{person}{Claire Foster}, \bibinfo{person}{Yeojin~Amy Ahn},
  \bibinfo{person}{Katarzyna Chawarska}, \bibinfo{person}{Brian Scassellati},
  {and} \bibinfo{person}{Frederick Shic}.} \bibinfo{year}{2016}\natexlab{}.
\newblock \showarticletitle{Emotional robot to examine different play patterns
  and affective responses of children with and without ASD}. In
  \bibinfo{booktitle}{\emph{2016 11th ACM/IEEE International Conference on
  Human-Robot Interaction (HRI)}}. IEEE, \bibinfo{pages}{19--26}.
\newblock


\bibitem[\protect\citeauthoryear{{Bradley Knox} and {Stone}}{{Bradley Knox} and
  {Stone}}{2008}]%
        {4640845}
\bibfield{author}{\bibinfo{person}{W. {Bradley Knox}} {and} \bibinfo{person}{P.
  {Stone}}.} \bibinfo{year}{2008}\natexlab{}.
\newblock \showarticletitle{TAMER: Training an Agent Manually via Evaluative
  Reinforcement}. In \bibinfo{booktitle}{\emph{2008 7th IEEE International
  Conference on Development and Learning}}. \bibinfo{pages}{292--297}.
\newblock
\showISSN{2161-9476}
\urldef\tempurl%
\url{https://doi.org/10.1109/DEVLRN.2008.4640845}
\showDOI{\tempurl}


\bibitem[\protect\citeauthoryear{Cakmak, DePalma, Thomaz, and Arriaga}{Cakmak
  et~al\mbox{.}}{2009}]%
        {cakmak2009effects}
\bibfield{author}{\bibinfo{person}{Maya Cakmak}, \bibinfo{person}{Nick
  DePalma}, \bibinfo{person}{Andrea~L Thomaz}, {and} \bibinfo{person}{Rosa
  Arriaga}.} \bibinfo{year}{2009}\natexlab{}.
\newblock \showarticletitle{Effects of social exploration mechanisms on robot
  learning}. In \bibinfo{booktitle}{\emph{RO-MAN 2009-The 18th IEEE
  International Symposium on Robot and Human Interactive Communication}}. IEEE,
  \bibinfo{pages}{128--134}.
\newblock


\bibitem[\protect\citeauthoryear{Chan, Hadfield-Menell, Srinivasa, and
  Dragan}{Chan et~al\mbox{.}}{2019}]%
        {chan2019assistive}
\bibfield{author}{\bibinfo{person}{Lawrence Chan}, \bibinfo{person}{Dylan
  Hadfield-Menell}, \bibinfo{person}{Siddhartha Srinivasa}, {and}
  \bibinfo{person}{Anca Dragan}.} \bibinfo{year}{2019}\natexlab{}.
\newblock \showarticletitle{The Assistive Multi-Armed Bandit}. In
  \bibinfo{booktitle}{\emph{2019 14th ACM/IEEE International Conference on
  Human-Robot Interaction (HRI)}}. IEEE, \bibinfo{pages}{354--363}.
\newblock


\bibitem[\protect\citeauthoryear{Cheung and Gardner}{Cheung and
  Gardner}{2015}]%
        {cheung2015way}
\bibfield{author}{\bibinfo{person}{Elaine~O Cheung} {and}
  \bibinfo{person}{Wendi~L Gardner}.} \bibinfo{year}{2015}\natexlab{}.
\newblock \showarticletitle{The way I make you feel: Social exclusion enhances
  the ability to manage others' emotions}.
\newblock \bibinfo{journal}{\emph{Journal of Experimental Social Psychology}}
  \bibinfo{volume}{60} (\bibinfo{year}{2015}), \bibinfo{pages}{59--75}.
\newblock


\bibitem[\protect\citeauthoryear{Dominey and Warneken}{Dominey and
  Warneken}{2011}]%
        {Dominey2011}
\bibfield{author}{\bibinfo{person}{P.~F. Dominey} {and} \bibinfo{person}{F.
  Warneken}.} \bibinfo{year}{2011}\natexlab{}.
\newblock \showarticletitle{The basis of shared intentions in human and robot
  cognition}.
\newblock \bibinfo{journal}{\emph{New ideas in psychology.}}
  \bibinfo{volume}{29}, \bibinfo{number}{3} (\bibinfo{year}{2011}).
\newblock


\bibitem[\protect\citeauthoryear{Fan, Tian, Qin, Li, and Liu}{Fan
  et~al\mbox{.}}{2018}]%
        {fan2018learning}
\bibfield{author}{\bibinfo{person}{Yang Fan}, \bibinfo{person}{Fei Tian},
  \bibinfo{person}{Tao Qin}, \bibinfo{person}{Xiang-Yang Li}, {and}
  \bibinfo{person}{Tie-Yan Liu}.} \bibinfo{year}{2018}\natexlab{}.
\newblock \showarticletitle{Learning to teach}.
\newblock \bibinfo{journal}{\emph{arXiv preprint arXiv:1805.03643}}
  (\bibinfo{year}{2018}).
\newblock


\bibitem[\protect\citeauthoryear{Fasola and Matari\'{c}}{Fasola and
  Matari\'{c}}{2013}]%
        {Fasola:2013:SAR:3109708.3109710}
\bibfield{author}{\bibinfo{person}{Juan Fasola} {and} \bibinfo{person}{Maja~J
  Matari\'{c}}.} \bibinfo{year}{2013}\natexlab{}.
\newblock \showarticletitle{A Socially Assistive Robot Exercise Coach for the
  Elderly}.
\newblock \bibinfo{journal}{\emph{J. Hum.-Robot Interact.}}
  \bibinfo{volume}{2}, \bibinfo{number}{2} (\bibinfo{date}{June}
  \bibinfo{year}{2013}), \bibinfo{pages}{3--32}.
\newblock
\showISSN{2163-0364}
\urldef\tempurl%
\url{https://doi.org/10.5898/JHRI.2.2.Fasola}
\showDOI{\tempurl}


\bibitem[\protect\citeauthoryear{Ferreira and Lef\`{e}vre}{Ferreira and
  Lef\`{e}vre}{2015}]%
        {Ferreira:2015:RBD:2794083.2794125}
\bibfield{author}{\bibinfo{person}{Emmanuel Ferreira} {and}
  \bibinfo{person}{Fabrice Lef\`{e}vre}.} \bibinfo{year}{2015}\natexlab{}.
\newblock \showarticletitle{Reinforcement-learning Based Dialogue System for
  Human-robot Interactions with Socially-inspired Rewards}.
\newblock \bibinfo{journal}{\emph{Comput. Speech Lang.}} \bibinfo{volume}{34},
  \bibinfo{number}{1} (\bibinfo{date}{Nov.} \bibinfo{year}{2015}),
  \bibinfo{pages}{256--274}.
\newblock
\showISSN{0885-2308}
\urldef\tempurl%
\url{https://doi.org/10.1016/j.csl.2015.03.007}
\showDOI{\tempurl}


\bibitem[\protect\citeauthoryear{Fitter and Kuchenbecker}{Fitter and
  Kuchenbecker}{2016}]%
        {Fitter-ICSR-16}
\bibfield{author}{\bibinfo{person}{Naomi~T Fitter} {and}
  \bibinfo{person}{Katherine~J Kuchenbecker}.} \bibinfo{year}{2016}\natexlab{}.
\newblock \showarticletitle{Designing and assessing expressive open-source
  faces for the Baxter robot}. In \bibinfo{booktitle}{\emph{Proceedings of the
  International Conference on Social Robotics (ICSR)}}.
\newblock


\bibitem[\protect\citeauthoryear{Griffith, Subramanian, Scholz, Isbell, and
  Thomaz}{Griffith et~al\mbox{.}}{2013}]%
        {griffith2013policy}
\bibfield{author}{\bibinfo{person}{Shane Griffith}, \bibinfo{person}{Kaushik
  Subramanian}, \bibinfo{person}{Jonathan Scholz}, \bibinfo{person}{Charles~L
  Isbell}, {and} \bibinfo{person}{Andrea~L Thomaz}.}
  \bibinfo{year}{2013}\natexlab{}.
\newblock \showarticletitle{Policy shaping: Integrating human feedback with
  reinforcement learning}. In \bibinfo{booktitle}{\emph{Advances in neural
  information processing systems}}. \bibinfo{pages}{2625--2633}.
\newblock


\bibitem[\protect\citeauthoryear{Hadfield-Menell, Russell, Abbeel, and
  Dragan}{Hadfield-Menell et~al\mbox{.}}{2016}]%
        {hadfield2016cooperative}
\bibfield{author}{\bibinfo{person}{Dylan Hadfield-Menell},
  \bibinfo{person}{Stuart~J Russell}, \bibinfo{person}{Pieter Abbeel}, {and}
  \bibinfo{person}{Anca Dragan}.} \bibinfo{year}{2016}\natexlab{}.
\newblock \showarticletitle{Cooperative inverse reinforcement learning}. In
  \bibinfo{booktitle}{\emph{Advances in neural information processing
  systems}}. \bibinfo{pages}{3909--3917}.
\newblock


\bibitem[\protect\citeauthoryear{Keskin}{Keskin}{2014}]%
        {keskin2014isn}
\bibfield{author}{\bibinfo{person}{Sevgi~Co{\c{s}}kun Keskin}.}
  \bibinfo{year}{2014}\natexlab{}.
\newblock \showarticletitle{From what isn’t empathy to empathic learning
  process}.
\newblock \bibinfo{journal}{\emph{Procedia-Social and Behavioral Sciences}}
  \bibinfo{volume}{116} (\bibinfo{year}{2014}), \bibinfo{pages}{4932--4938}.
\newblock


\bibitem[\protect\citeauthoryear{Kim, Berkovits, Bernier, Leyzberg, Shic, Paul,
  and Scassellati}{Kim et~al\mbox{.}}{2013}]%
        {kim2013social}
\bibfield{author}{\bibinfo{person}{Elizabeth~S Kim}, \bibinfo{person}{Lauren~D
  Berkovits}, \bibinfo{person}{Emily~P Bernier}, \bibinfo{person}{Dan
  Leyzberg}, \bibinfo{person}{Frederick Shic}, \bibinfo{person}{Rhea Paul},
  {and} \bibinfo{person}{Brian Scassellati}.} \bibinfo{year}{2013}\natexlab{}.
\newblock \showarticletitle{Social robots as embedded reinforcers of social
  behavior in children with autism}.
\newblock \bibinfo{journal}{\emph{Journal of autism and developmental
  disorders}} \bibinfo{volume}{43}, \bibinfo{number}{5} (\bibinfo{year}{2013}),
  \bibinfo{pages}{1038--1049}.
\newblock


\bibitem[\protect\citeauthoryear{Knox and Stone}{Knox and Stone}{2009}]%
        {Knox:2009:ISA:1597735.1597738}
\bibfield{author}{\bibinfo{person}{W.~Bradley Knox} {and}
  \bibinfo{person}{Peter Stone}.} \bibinfo{year}{2009}\natexlab{}.
\newblock \showarticletitle{Interactively Shaping Agents via Human
  Reinforcement: The TAMER Framework}. In \bibinfo{booktitle}{\emph{Proceedings
  of the Fifth International Conference on Knowledge Capture}}
  \emph{(\bibinfo{series}{K-CAP '09})}. \bibinfo{publisher}{ACM},
  \bibinfo{address}{New York, NY, USA}, \bibinfo{pages}{9--16}.
\newblock
\showISBNx{978-1-60558-658-8}
\urldef\tempurl%
\url{https://doi.org/10.1145/1597735.1597738}
\showDOI{\tempurl}


\bibitem[\protect\citeauthoryear{Koenig, Takayama, and Mataric}{Koenig
  et~al\mbox{.}}{2010}]%
        {Koenig2010}
\bibfield{author}{\bibinfo{person}{N. Koenig}, \bibinfo{person}{L. Takayama},
  {and} \bibinfo{person}{M. Mataric}.} \bibinfo{year}{2010}\natexlab{}.
\newblock \showarticletitle{Communication and knowledge sharing in
  human–robot interaction and learning from demonstration}.
\newblock \bibinfo{journal}{\emph{IEEE transactions on neural networks}}
  \bibinfo{volume}{23}, \bibinfo{number}{8-9} (\bibinfo{year}{2010}).
\newblock
\showISSN{1045-9227}


\bibitem[\protect\citeauthoryear{Konok}{Konok}{2018}]%
        {Konok2018}
\bibfield{author}{\bibinfo{person}{Veronika Konok}.}
  \bibinfo{year}{2018}\natexlab{}.
\newblock \showarticletitle{Should we love robots?–The most liked qualities
  of companion dogs and how they can be implemented in social robots.}
\newblock \bibinfo{journal}{\emph{Computers in Human Behaviour}}
  \bibinfo{volume}{80} (\bibinfo{year}{2018}), \bibinfo{pages}{132--142}.
\newblock


\bibitem[\protect\citeauthoryear{Lazarescu}{Lazarescu}{2013}]%
        {Mihaela2013}
\bibfield{author}{\bibinfo{person}{Mihaela~Paisi. Lazarescu}.}
  \bibinfo{year}{2013}\natexlab{}.
\newblock \showarticletitle{The Structure and Dynamics of the Teacher's
  Empathic Behavior}.
\newblock \bibinfo{journal}{\emph{Procedia-Social and Behavioral Sciences}}
  \bibinfo{volume}{78} (\bibinfo{year}{2013}), \bibinfo{pages}{511--515}.
\newblock


\bibitem[\protect\citeauthoryear{Lee, Lau, Kiesler, and Chiu}{Lee
  et~al\mbox{.}}{2005}]%
        {lee2005human}
\bibfield{author}{\bibinfo{person}{Sau-lai Lee}, \bibinfo{person}{Ivy Yee-man
  Lau}, \bibinfo{person}{Sara Kiesler}, {and} \bibinfo{person}{Chi-Yue Chiu}.}
  \bibinfo{year}{2005}\natexlab{}.
\newblock \showarticletitle{Human mental models of humanoid robots}. In
  \bibinfo{booktitle}{\emph{Proceedings of the 2005 IEEE international
  conference on robotics and automation}}. IEEE, \bibinfo{pages}{2767--2772}.
\newblock


\bibitem[\protect\citeauthoryear{Leite, Castellano, Pereira, Martinho, and
  Paiva}{Leite et~al\mbox{.}}{2012a}]%
        {leite2012long}
\bibfield{author}{\bibinfo{person}{Iolanda Leite}, \bibinfo{person}{Ginevra
  Castellano}, \bibinfo{person}{Andr{\'e} Pereira}, \bibinfo{person}{Carlos
  Martinho}, {and} \bibinfo{person}{Ana Paiva}.}
  \bibinfo{year}{2012}\natexlab{a}.
\newblock \showarticletitle{Long-term interactions with empathic robots:
  Evaluating perceived support in children}. In
  \bibinfo{booktitle}{\emph{International Conference on Social Robotics}}.
  Springer, \bibinfo{pages}{298--307}.
\newblock


\bibitem[\protect\citeauthoryear{Leite, Pereira, Mascarenhas, Martinho, Prada,
  and Paiva}{Leite et~al\mbox{.}}{2012b}]%
        {article2012}
\bibfield{author}{\bibinfo{person}{Iolanda Leite}, \bibinfo{person}{André
  Pereira}, \bibinfo{person}{Samuel Mascarenhas}, \bibinfo{person}{Carlos
  Martinho}, \bibinfo{person}{Rui Prada}, {and} \bibinfo{person}{Ana Paiva}.}
  \bibinfo{year}{2012}\natexlab{b}.
\newblock \showarticletitle{The influence of empathy in human–robot
  relations}.
\newblock \bibinfo{journal}{\emph{International Journal of Human-Computer
  Studies}}  \bibinfo{volume}{71} (\bibinfo{date}{01} \bibinfo{year}{2012}).
\newblock
\urldef\tempurl%
\url{https://doi.org/10.1016/j.ijhcs.2012.09.005}
\showDOI{\tempurl}


\bibitem[\protect\citeauthoryear{Litoiu and Scassellati}{Litoiu and
  Scassellati}{2015}]%
        {litoiu2015robotic}
\bibfield{author}{\bibinfo{person}{Alexandru Litoiu} {and}
  \bibinfo{person}{Brian Scassellati}.} \bibinfo{year}{2015}\natexlab{}.
\newblock \showarticletitle{Robotic coaching of complex physical skills}. In
  \bibinfo{booktitle}{\emph{Proceedings of the Tenth Annual ACM/IEEE
  International Conference on Human-Robot Interaction Extended Abstracts}}.
  ACM, \bibinfo{pages}{211--212}.
\newblock


\bibitem[\protect\citeauthoryear{Lockwood}{Lockwood}{2016}]%
        {lockwood2016anatomy}
\bibfield{author}{\bibinfo{person}{Patricia~L Lockwood}.}
  \bibinfo{year}{2016}\natexlab{}.
\newblock \showarticletitle{The anatomy of empathy: Vicarious experience and
  disorders of social cognition}.
\newblock \bibinfo{journal}{\emph{Behavioural brain research}}
  \bibinfo{volume}{311} (\bibinfo{year}{2016}), \bibinfo{pages}{255--266}.
\newblock


\bibitem[\protect\citeauthoryear{Lopes, Melo, and Montesano}{Lopes
  et~al\mbox{.}}{2009}]%
        {Lopes-ECML-09}
\bibfield{author}{\bibinfo{person}{Manuel Lopes}, \bibinfo{person}{Francisco
  Melo}, {and} \bibinfo{person}{Luis Montesano}.}
  \bibinfo{year}{2009}\natexlab{}.
\newblock \showarticletitle{Active learning for reward estimation in inverse
  reinforcement learning}. In \bibinfo{booktitle}{\emph{Proceedings of the
  Joint European Conference on Machine Learning (ECML)}}.
\newblock


\bibitem[\protect\citeauthoryear{Lopes, Roy, Oudeyer, and Clement}{Lopes
  et~al\mbox{.}}{2013}]%
        {Lopes2013multi}
\bibfield{author}{\bibinfo{person}{Manuel Lopes}, \bibinfo{person}{Didier Roy},
  \bibinfo{person}{Pierre-Yves Oudeyer}, {and} \bibinfo{person}{Benjamin
  Clement}.} \bibinfo{year}{2013}\natexlab{}.
\newblock \showarticletitle{Multi-armed bandits for intelligent tutoring
  systems}.
\newblock \bibinfo{journal}{\emph{arXiv preprint arXiv:1310.3174}}
  (\bibinfo{year}{2013}).
\newblock


\bibitem[\protect\citeauthoryear{Nikolaidis and Shah}{Nikolaidis and
  Shah}{2013}]%
        {nikolaidis2013human}
\bibfield{author}{\bibinfo{person}{Stefanos Nikolaidis} {and}
  \bibinfo{person}{Julie Shah}.} \bibinfo{year}{2013}\natexlab{}.
\newblock \showarticletitle{Human-robot cross-training: computational
  formulation, modeling and evaluation of a human team training strategy}. In
  \bibinfo{booktitle}{\emph{Proceedings of the 8th ACM/IEEE international
  conference on Human-robot interaction}}. IEEE Press, \bibinfo{pages}{33--40}.
\newblock


\bibitem[\protect\citeauthoryear{Nunez, Matsuda, Hirokawa, and Suzuki}{Nunez
  et~al\mbox{.}}{2015}]%
        {nunez2015humanoid}
\bibfield{author}{\bibinfo{person}{Eleuda Nunez}, \bibinfo{person}{Soichiro
  Matsuda}, \bibinfo{person}{Masakazu Hirokawa}, {and} \bibinfo{person}{Kenji
  Suzuki}.} \bibinfo{year}{2015}\natexlab{}.
\newblock \showarticletitle{Humanoid robot assisted training for facial
  expressions recognition based on affective feedback}. In
  \bibinfo{booktitle}{\emph{International Conference on Social Robotics}}.
  Springer, \bibinfo{pages}{492--501}.
\newblock


\bibitem[\protect\citeauthoryear{Park, Rosenberg-Kima, Rosenberg, Gordon, and
  Breazeal}{Park et~al\mbox{.}}{2017}]%
        {park2017growing}
\bibfield{author}{\bibinfo{person}{Hae~Won Park}, \bibinfo{person}{Rinat
  Rosenberg-Kima}, \bibinfo{person}{Maor Rosenberg}, \bibinfo{person}{Goren
  Gordon}, {and} \bibinfo{person}{Cynthia Breazeal}.}
  \bibinfo{year}{2017}\natexlab{}.
\newblock \showarticletitle{Growing growth mindset with a social robot peer}.
  In \bibinfo{booktitle}{\emph{2017 12th ACM/IEEE International Conference on
  Human-Robot Interaction (HRI}}. IEEE, \bibinfo{pages}{137--145}.
\newblock


\bibitem[\protect\citeauthoryear{Powell and Roberts}{Powell and
  Roberts}{2016}]%
        {Powell2016}
\bibfield{author}{\bibinfo{person}{Philip Powell} {and}
  \bibinfo{person}{Jennifer Roberts}.} \bibinfo{year}{2016}\natexlab{}.
\newblock \showarticletitle{Situational determinants of cognitive, affective,
  and compassionate empathy in naturalistic digital interactions}.
\newblock \bibinfo{journal}{\emph{Computers in Human Behaviour}}
  \bibinfo{volume}{68} (\bibinfo{date}{11} \bibinfo{year}{2016}).
\newblock
\urldef\tempurl%
\url{https://doi.org/10.1016/j.chb.2016.11.024}
\showDOI{\tempurl}


\bibitem[\protect\citeauthoryear{Prozesky}{Prozesky}{2000}]%
        {prozesky2000teaching}
\bibfield{author}{\bibinfo{person}{Detlef~R Prozesky}.}
  \bibinfo{year}{2000}\natexlab{}.
\newblock \showarticletitle{Teaching and learning}.
\newblock \bibinfo{journal}{\emph{Community eye health}} \bibinfo{volume}{13},
  \bibinfo{number}{36} (\bibinfo{year}{2000}), \bibinfo{pages}{60}.
\newblock


\bibitem[\protect\citeauthoryear{Ramachandran and Scassellati}{Ramachandran and
  Scassellati}{2015}]%
        {ramachandran2015developing}
\bibfield{author}{\bibinfo{person}{Aditi Ramachandran} {and}
  \bibinfo{person}{Brian Scassellati}.} \bibinfo{year}{2015}\natexlab{}.
\newblock \showarticletitle{Developing adaptive social robot tutors for
  children}. In \bibinfo{booktitle}{\emph{2015 AAAI Fall Symposium Series}}.
\newblock


\bibitem[\protect\citeauthoryear{Roy, Crick, Kieson, and Abramson}{Roy
  et~al\mbox{.}}{2018a}]%
        {roy2018reinforcement}
\bibfield{author}{\bibinfo{person}{Sayanti Roy}, \bibinfo{person}{Christopher
  Crick}, \bibinfo{person}{Emily Kieson}, {and} \bibinfo{person}{Charles
  Abramson}.} \bibinfo{year}{2018}\natexlab{a}.
\newblock \showarticletitle{A reinforcement learning model for robots as
  teachers}. In \bibinfo{booktitle}{\emph{2018 27th IEEE International
  Symposium on Robot and Human Interactive Communication (RO-MAN)}}. IEEE,
  \bibinfo{pages}{294--299}.
\newblock


\bibitem[\protect\citeauthoryear{Roy, Kieson, Abramson, and Crick}{Roy
  et~al\mbox{.}}{2018b}]%
        {roy2018using}
\bibfield{author}{\bibinfo{person}{Sayanti Roy}, \bibinfo{person}{Emily
  Kieson}, \bibinfo{person}{Charles Abramson}, {and}
  \bibinfo{person}{Christopher Crick}.} \bibinfo{year}{2018}\natexlab{b}.
\newblock \showarticletitle{Using human reinforcement learning models to
  improve robots as teachers}. In \bibinfo{booktitle}{\emph{Companion of the
  2018 ACM/IEEE International Conference on Human-Robot Interaction}}. ACM,
  \bibinfo{pages}{225--226}.
\newblock


\bibitem[\protect\citeauthoryear{Roy, Kieson, Abramson, and Crick}{Roy
  et~al\mbox{.}}{2019a}]%
        {royeffective}
\bibfield{author}{\bibinfo{person}{Sayanti Roy}, \bibinfo{person}{Emily
  Kieson}, \bibinfo{person}{Charles Abramson}, {and}
  \bibinfo{person}{Christopher Crick}.} \bibinfo{year}{2019}\natexlab{a}.
\newblock \showarticletitle{Effective robot-human skill transfer via mutual
  reinforcement learning}.
\newblock  (\bibinfo{year}{2019}).
\newblock


\bibitem[\protect\citeauthoryear{Roy, Kieson, Abramson, and Crick}{Roy
  et~al\mbox{.}}{2019b}]%
        {roy2019mutual}
\bibfield{author}{\bibinfo{person}{Sayanti Roy}, \bibinfo{person}{Emily
  Kieson}, \bibinfo{person}{Charles Abramson}, {and}
  \bibinfo{person}{Christopher Crick}.} \bibinfo{year}{2019}\natexlab{b}.
\newblock \showarticletitle{Mutual Reinforcement Learning with Robot Trainers}.
  In \bibinfo{booktitle}{\emph{2019 14th ACM/IEEE International Conference on
  Human-Robot Interaction (HRI)}}. IEEE, \bibinfo{pages}{572--573}.
\newblock


\bibitem[\protect\citeauthoryear{Salter, Dautenhahn, and Boekhorst}{Salter
  et~al\mbox{.}}{2006}]%
        {Salter2006}
\bibfield{author}{\bibinfo{person}{Tamie Salter}, \bibinfo{person}{Kerstin
  Dautenhahn}, {and} \bibinfo{person}{Rene Boekhorst}.}
  \bibinfo{year}{2006}\natexlab{}.
\newblock \showarticletitle{Learning about natural human-robot interaction
  styles}.
\newblock \bibinfo{journal}{\emph{Robotics and Autonomous Systems}}
  \bibinfo{volume}{54(2)} (\bibinfo{date}{02} \bibinfo{year}{2006}),
  \bibinfo{pages}{127--134}.
\newblock
\urldef\tempurl%
\url{https://doi.org/10.1016/j.robot.2005.09.022}
\showDOI{\tempurl}


\bibitem[\protect\citeauthoryear{Saupp{\'e} and Mutlu}{Saupp{\'e} and
  Mutlu}{2015}]%
        {Sauppe:2015:ETT:2825762.2825778}
\bibfield{author}{\bibinfo{person}{Allison Saupp{\'e}} {and}
  \bibinfo{person}{Bilge Mutlu}.} \bibinfo{year}{2015}\natexlab{}.
\newblock \showarticletitle{Effective Task Training Strategies for Human and
  Robot Instructors}.
\newblock \bibinfo{journal}{\emph{Auton. Robots}} \bibinfo{volume}{39},
  \bibinfo{number}{3} (\bibinfo{date}{Oct.} \bibinfo{year}{2015}),
  \bibinfo{pages}{313--329}.
\newblock
\showISSN{0929-5593}
\urldef\tempurl%
\url{https://doi.org/10.1007/s10514-015-9461-0}
\showDOI{\tempurl}


\bibitem[\protect\citeauthoryear{Scassellati}{Scassellati}{2001}]%
        {scassellati2001foundations}
\bibfield{author}{\bibinfo{person}{Brian~M Scassellati}.}
  \bibinfo{year}{2001}\natexlab{}.
\newblock \emph{\bibinfo{title}{Foundations for a Theory of Mind for a Humanoid
  Robot}}.
\newblock \bibinfo{thesistype}{Ph.D. Dissertation}.
  \bibinfo{school}{Massachusetts Institute of Technology}.
\newblock


\bibitem[\protect\citeauthoryear{Shannon}{Shannon}{1948}]%
        {shannon1948mathematical}
\bibfield{author}{\bibinfo{person}{Claude~Elwood Shannon}.}
  \bibinfo{year}{1948}\natexlab{}.
\newblock \showarticletitle{A mathematical theory of communication}.
\newblock \bibinfo{journal}{\emph{Bell system technical journal}}
  \bibinfo{volume}{27}, \bibinfo{number}{3} (\bibinfo{year}{1948}),
  \bibinfo{pages}{379--423}.
\newblock


\bibitem[\protect\citeauthoryear{Simpson}{Simpson}{1972}]%
        {simpson1972classification}
\bibfield{author}{\bibinfo{person}{E Simpson}.}
  \bibinfo{year}{1972}\natexlab{}.
\newblock \showarticletitle{The classification objectives in the psychomotor
  domain--Washington}.
\newblock \bibinfo{journal}{\emph{Gryphon House}} (\bibinfo{year}{1972}).
\newblock


\bibitem[\protect\citeauthoryear{Simpson}{Simpson}{1966}]%
        {simpson1966classification}
\bibfield{author}{\bibinfo{person}{Elizabeth~J Simpson}.}
  \bibinfo{year}{1966}\natexlab{}.
\newblock \showarticletitle{The Classification of Educational Objectives,
  Psychomotor Domain.}
\newblock  (\bibinfo{year}{1966}).
\newblock


\bibitem[\protect\citeauthoryear{Spaulding, Chen, Ali, Kulinski, and
  Breazeal}{Spaulding et~al\mbox{.}}{2018}]%
        {spaulding2018social}
\bibfield{author}{\bibinfo{person}{Samuel Spaulding}, \bibinfo{person}{Huili
  Chen}, \bibinfo{person}{Safinah Ali}, \bibinfo{person}{Michael Kulinski},
  {and} \bibinfo{person}{Cynthia Breazeal}.} \bibinfo{year}{2018}\natexlab{}.
\newblock \showarticletitle{A Social Robot System for Modeling Children's Word
  Pronunciation: Socially Interactive Agents Track}. In
  \bibinfo{booktitle}{\emph{Proceedings of the 17th International Conference on
  Autonomous Agents and MultiAgent Systems}}. International Foundation for
  Autonomous Agents and Multiagent Systems, \bibinfo{pages}{1658--1666}.
\newblock


\bibitem[\protect\citeauthoryear{Stojanov, Trajkovski, and Kulakov}{Stojanov
  et~al\mbox{.}}{2006}]%
        {Stojanov2006}
\bibfield{author}{\bibinfo{person}{Georgi Stojanov}, \bibinfo{person}{Goran
  Trajkovski}, {and} \bibinfo{person}{Andrea Kulakov}.}
  \bibinfo{year}{2006}\natexlab{}.
\newblock \showarticletitle{Interactivism in artificial intelligence (AI) and
  intelligent robotics}.
\newblock \bibinfo{journal}{\emph{New ideas in psychology.}}
  \bibinfo{volume}{24}, \bibinfo{number}{2} (\bibinfo{year}{2006}).
\newblock
\showISSN{0732-118X}


\bibitem[\protect\citeauthoryear{Sutton and Barto}{Sutton and Barto}{2018}]%
        {sutton2018reinforcement}
\bibfield{author}{\bibinfo{person}{Richard~S Sutton} {and}
  \bibinfo{person}{Andrew~G Barto}.} \bibinfo{year}{2018}\natexlab{}.
\newblock \bibinfo{booktitle}{\emph{Reinforcement learning: An introduction}}.
\newblock \bibinfo{publisher}{MIT press}.
\newblock


\bibitem[\protect\citeauthoryear{Tabrez, Agrawal, and Hayes}{Tabrez
  et~al\mbox{.}}{2019}]%
        {tabrez2019explanation}
\bibfield{author}{\bibinfo{person}{Aaquib Tabrez}, \bibinfo{person}{Shivendra
  Agrawal}, {and} \bibinfo{person}{Bradley Hayes}.}
  \bibinfo{year}{2019}\natexlab{}.
\newblock \showarticletitle{Explanation-based reward coaching to improve human
  performance via reinforcement learning}. In \bibinfo{booktitle}{\emph{2019
  14th ACM/IEEE International Conference on Human-Robot Interaction (HRI)}}.
  IEEE, \bibinfo{pages}{249--257}.
\newblock


\bibitem[\protect\citeauthoryear{Yi, Sun, Li, and Li}{Yi et~al\mbox{.}}{2018}]%
        {yi2018automatic}
\bibfield{author}{\bibinfo{person}{Xiaoyuan Yi}, \bibinfo{person}{Maosong Sun},
  \bibinfo{person}{Ruoyu Li}, {and} \bibinfo{person}{Wenhao Li}.}
  \bibinfo{year}{2018}\natexlab{}.
\newblock \showarticletitle{Automatic Poetry Generation with Mutual
  Reinforcement Learning}. In \bibinfo{booktitle}{\emph{Proceedings of the 2018
  Conference on Empirical Methods in Natural Language Processing}}.
  \bibinfo{pages}{3143--3153}.
\newblock


\bibitem[\protect\citeauthoryear{Yin, Billard, and Paiva}{Yin
  et~al\mbox{.}}{2015}]%
        {yin2015bidirectional}
\bibfield{author}{\bibinfo{person}{Hang Yin}, \bibinfo{person}{Aude Billard},
  {and} \bibinfo{person}{Ana Paiva}.} \bibinfo{year}{2015}\natexlab{}.
\newblock \showarticletitle{Bidirectional learning of handwriting skill in
  human-robot interaction}. In \bibinfo{booktitle}{\emph{Proceedings of the
  Tenth Annual ACM/IEEE International Conference on Human-Robot Interaction
  Extended Abstracts}}. ACM, \bibinfo{pages}{243--244}.
\newblock


\end{thebibliography}

\end{document}